\RequirePackage{fix-cm}
\documentclass[twocolumn]{svjour3}          % twocolumn
\smartqed  % flush right qed marks, e.g. at end of proof
\usepackage{graphicx}
\usepackage{pifont}% http://ctan.org/pkg/pifont
\newcommand{\cmark}{\ding{51}}%
\newcommand{\xmark}{\ding{55}}%
\usepackage{makecell}
\usepackage{multirow}
\usepackage{booktabs}
\usepackage{enumerate}
\usepackage{cite}

\newcommand{\PreserveBackslash}[1]{\let\temp=\\#1\let\\=\temp}
\newcolumntype{C}[1]{>{\PreserveBackslash\centering}p{#1}}
\newcolumntype{R}[1]{>{\PreserveBackslash\raggedleft}p{#1}}
\newcolumntype{L}[1]{>{\PreserveBackslash\raggedright}p{#1}}

\newcommand{\tabincell}[2]{\begin{tabular}{@{}#1@{}}#2\end{tabular}}

\usepackage[breaklinks=true,letterpaper=true,colorlinks,citecolor=blue,bookmarks=false]{hyperref}
\newcommand{\eg}{\emph{e.g.}}
\newcommand{\ie}{\emph{i.e.}}

\journalname{International Journal of Computer Vision}
\begin{document}

\title{LaSOT: A High-quality Large-scale Single Object Tracking Benchmark
	\thanks{$^{*}$ H. Fan and H. Bai make equal contribution to this work.}
}
% \subtitle{Do you have a subtitle?\\ If so, write it here}


\author{Heng Fan$^{*}$ \and Hexin Bai$^{*}$ \and Liting Lin \and Fan Yang \and Peng Chu  \and Ge Deng \and Sijia Yu \and Harshit \and Mingzhen Huang \and Juehuan Liu \and Yong Xu \and Chunyuan Liao \and Lin Yuan \and Haibin Ling}


%\institute{
%			H. Fan, M. Huang, Harshit, H. Ling \at
%			Department of Computer Science, Stony Brook University, Stony Brook, USA 
%			\and
%			H. Bai, F. Yang, P. Chu, G. Deng, S. Yu \at
%			Department of Computer and Information Sciences, Temple University, Philadelphia, USA 
%			\and
%			L. Lin, Y. Xu \at
%			School of Computer Science \& Engineering, South China Univ. of Tech., Guangzhou, China \\
%			Peng Cheng Laboratory, Shenzhen, China
%			\and
%			C. Liao \at
%			Meitu HiScene Lab, HiScene Information Technologies, Shanghai, China 
%			\and
%			L. Yuan \at
%			Amazon Web Services, Palo Alto, USA
%			\and
%			E-mail: \{hefan, hling\}@cs.stonybrook.edu
%}
\institute{
	H. Fan, Harshit, M. Huang, H. Ling \at
	Stony Brook University, Stony Brook, USA 
	\and
	H. Bai, F. Yang, P. Chu, G. Deng, S. Yu, J. Liu \at
	Temple University, Philadelphia, USA 
	\and
	L. Lin, Y. Xu \at
	South China University of Technology, Guangzhou, China \\
	Peng Cheng Laboratory, Shenzhen, China
	\and
	C. Liao \at
	HiScene Information Technologies, Shanghai, China 
	\and
	L. Yuan \at
	Amazon Web Services, Palo Alto, USA
	\and
	E-mail: \{hefan, hling\}@cs.stonybrook.edu
}
\date{Received: date / Accepted: date}

\maketitle

\begin{abstract}
Despite great recent advances in visual tracking, its further development, including both algorithm design and evaluation, is limited due to lack of dedicated large-scale benchmarks. To address this problem, we present {\bf LaSOT}, a high-quality {\bf La}rge-scale {\bf S}ingle {\bf O}bject {\bf T}racking benchmark. LaSOT contains a diverse selection of 85 object classes, and offers 1,550 totaling more than 3.87 million frames. Each video frame is carefully and manually annotated with a bounding box. This makes LaSOT, to our knowledge, the largest densely annotated tracking benchmark. Our goal in releasing LaSOT is to provide a dedicated high quality platform for both training and evaluation of trackers. The average video length of LaSOT is around 2,500 frames, where each video contains various challenge factors that exist in real world video footage,such as the targets disappearing and re-appearing. These longer video lengths allow for the assessment of long-term trackers. To take advantage of the close connection between visual appearance and natural language, we provide language specification for each video in LaSOT. We believe such additions will allow for future research to use linguistic features to improve tracking. Two protocols, \emph{full-overlap} and \emph{one-shot}, are designated for flexible assessment of trackers. We extensively evaluate 48 baseline trackers on LaSOT with in-depth analysis, and results reveal that there still exists significant room for improvement. The complete benchmark, tracking results as well as analysis are available at \url{http://vision.cs.stonybrook.edu/~lasot/}.

\end{abstract}

%%%%%%%%%%%%%%%%%%%%%%%%%%%%%%%%%%%%%%%%%%%%%%%%%%%%%%%%%%%%%%%%%%%%%%%%%%%%%%%%%%%%%%%%%%%%%%
% Introduction
%%%%%%%%%%%%%%%%%%%%%%%%%%%%%%%%%%%%%%%%%%%%%%%%%%%%%%%%%%%%%%%%%%%%%%%%%%%%%%%%%%%%%%%%%%%%%%

\section{Introduction}

Visual object tracking plays a crucial role in computer vision and has a wide range of applications including intelligent vehicles, robotics, human-machine interaction, and surveillance~\cite{li2013survey,smeulders2014visual,yilmaz2006object}. Among various types of tracking problems, a popular and fundamental one is the so-called model-free generic object tracking, which is the focus of this paper. Briefly speaking, given the target bounding box in the initial frame, the goal of tracking is to locate the target in a video sequentially. 

In recent years, considerable progress has been made in improving tracking performance. Visual tracking benchmarks have been playing a key role in providing fair comparison and evaluation of different trackers, advancing the research frontier of visual tracking significantly. However, current benchmarks have limited further development of tracking in the deep learning era, as well as more authentic performance evaluation in real world scenarios, due to the following reasons:

\begin{figure}[!t]
	\centering
	\includegraphics[width=1\linewidth]{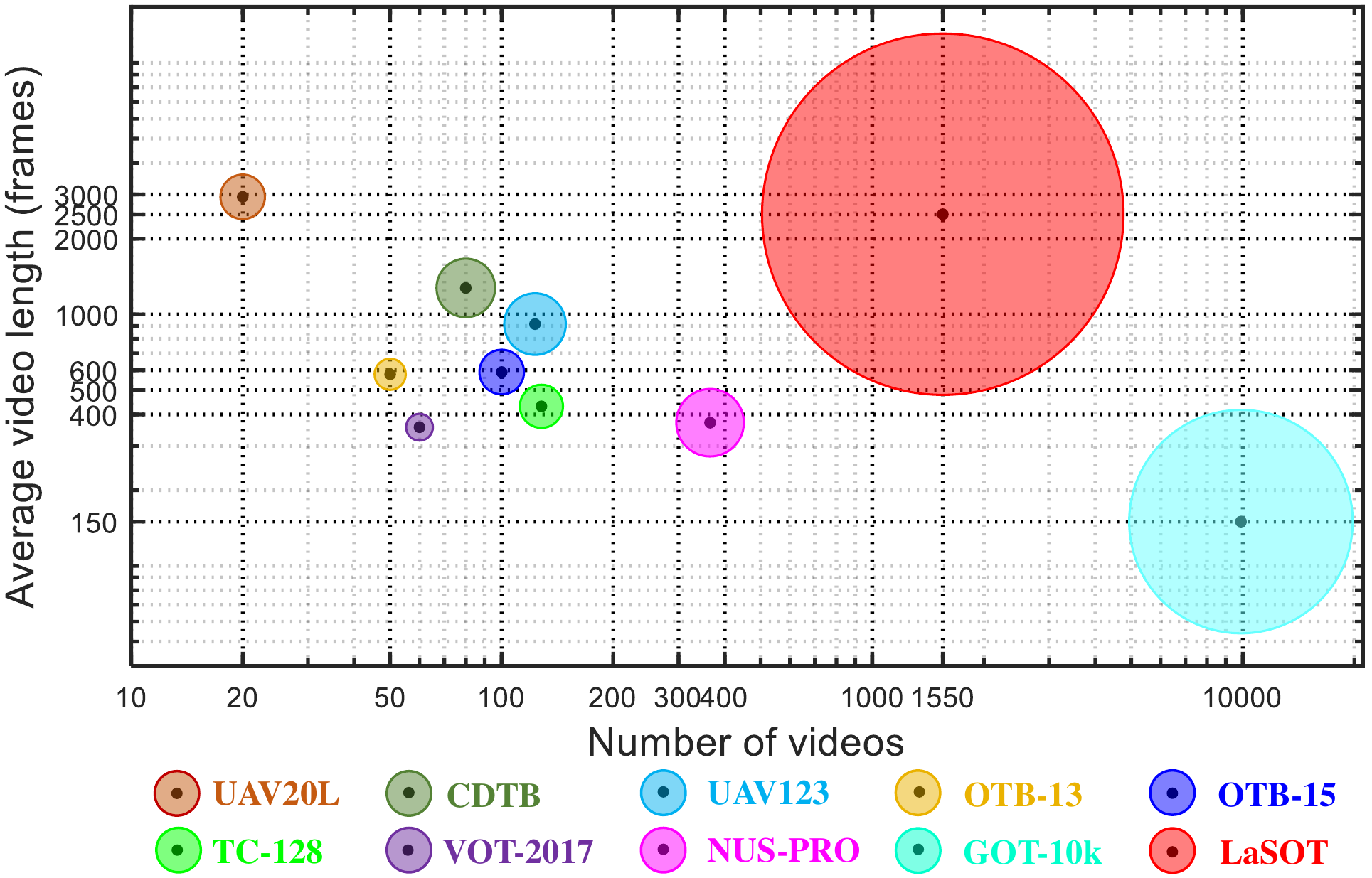}\\
	\caption{Summary of existing benchmarks with precise dense (per frame) annotations using log-log scale, containing OTB-13~\cite{wu2013online}, OTB-15~\cite{wu2015object}, TC-128~\cite{liang2015encoding}, NUS-PRO~\cite{li2016nus}, UAV123~\cite{mueller2016benchmark}, UAV20L~\cite{mueller2016benchmark}, CDTB~\cite{lukezic2019cdtb}, VOT-2017~\cite{kristan2017visual}, GOT-10k~\cite{huang2018got} and LaSOT. The circle diameter is in proportion to the number of frames of a benchmark. The proposed LaSOT is {\it larger} than all other benchmarks with more than 3.87M frames, and focused on {\it long-term} tracking with average video length of around 2,500 frames. Best viewed in color.}
	\label{fig:benchmark-scale}
\end{figure}

%\vspace{.1em}
\noindent{\bf{Small-scale.}} Motivated by the success of deep learning~\cite{krizhevsky2012imagenet,he2016deep,simonyan2015very}, deep feature representation has been widely adopted for target appearance modeling in tracking and has achieved significant improvements. To learn a robust deep representation, a dedicated \emph{large-scale} tracking benchmark is needed. However, most existing datasets contain less than 400 videos (see Figure~\ref{fig:benchmark-scale}), which makes it hard to learn a \emph{tracking-specific} deep representation. Consequently, researchers in the tracking community have been forced to leverage either the pre-trained models (\eg,~\cite{krizhevsky2012imagenet},~\cite{simonyan2015very} and~\cite{he2016deep}) from ImageNet~\cite{deng2009imagenet} for deep feature extraction or the sequences from video object detection (\eg,~\cite{russakovsky2015imagenet} and~\cite{real2017youtube}) for deep feature learning, which may result in suboptimal tracking performance owing to intrinsic differences between different tasks~\cite{yosinski2014transferable}. Extensive evaluation on \emph{large-scale} benchmark is needed to reliably demonstrate performance and generality of trackers.

\noindent{\bf{Lack of high-quality dense annotations.}} Accurate and dense (\ie, per-frame) annotations are crucial to visual object tracking for several reasons. They ensure more accurate and reliable evaluations and more fair comparisons for different trackers, offer desired training samples for developing tracking algorithms, and provide rich motion information and temporal context in videos. It is worth noting that there have been benchmarks proposed recently built towards large-scale and long-term tracking, such as~\cite{muller2018trackingnet} and~\cite{valmadre2018long}. However, their annotations are either semi-automatic (\eg, generated by a tracking algorithm) or sparse (\eg, labeled every 30 frames), limiting their usability.

\noindent{\bf{Short-term tracking.}} In order to be deployed in practical application, a tracking algorithm should be able to work well in a long sequence where the target object may frequently leave and enter the view. However, most current tracking datasets contain shorter length videos making them \emph{short-term} benchmarks. As shown in Figure~\ref{fig:benchmark-scale}, the average video length of these benchmarks is less then 600 frames (\ie, 20 seconds for 30\emph{fps} video rate). In addition, in these \emph{short-term} benchmarks, the target objects almost always appear in the video view. As a consequence, the evaluations on such \emph{short-term} benchmarks may not reflect the performance of an algorithm in the real world, and thus restrict applications.

\begin{figure}[!t]
	\centering
	\includegraphics[width=\linewidth]{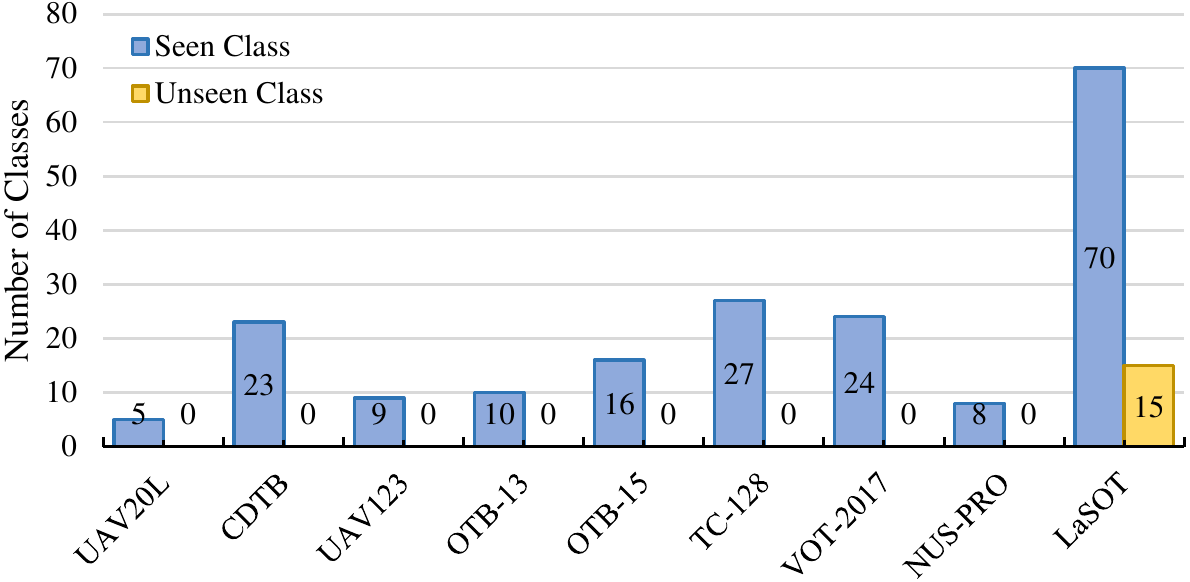}
	\caption{Comparison of number of classes for evaluation in densely annotated benchmarks, containing UAV20L~\cite{mueller2016benchmark}, CDTB~\cite{lukezic2019cdtb}, UAV123~\cite{mueller2016benchmark}, OTB-13~\cite{wu2013online}, OTB-15~\cite{wu2015object}, TC-128~\cite{liang2015encoding}, VOT-2017~\cite{kristan2017visual}, NUS-PRO~\cite{li2016nus} and LaSOT. We observe that the proposed LaSOT contains the most seen object categories, containing 70 different ones. Moreover, 15 extra unseen object classes are provided for new one-shot evaluation. Note that, GOT-10k~\cite{huang2018got} is not included for comparison because its method to count object classes is different from existing benchmarks.}
	\label{fig:figurenumberclass}
\end{figure}

\noindent
{\bf Limited number of object categories.} To assess the performance of tracking algorithms in the real world, it is necessary to utilize a diverse set of object categories for evaluation. However, most existing benchmarks contain less than 30 object categories for evaluation (see Figure~\ref{fig:figurenumberclass}). In addition, these benchmarks do not provide any unseen object classes in evaluation, which makes it difficult to fully evaluate the tracking performance in real applications. We note that the recent GOT-10k~\cite{huang2018got} tackles this problem by introducing a large set of object classes for tracking.

\noindent{\bf{Category bias.}} A robust tracker should demonstrate stable performance in locating arbitrary targets regardless of their categories, which requires that \emph{category bias} (or \emph{class imbalance}) should be eliminated in training and/or evaluating tracking algorithms. Despite this, most current tracking benchmarks usually consist of a few object classes (see Table~\ref{tab:DMA_comparison}). The GOT-10k~\cite{huang2018got} alleviates the problem of category bias to some extent by introducing diverse categories. However, categories are not rigorously balanced as the number of videos varies a lot across different categories.

\noindent{\bf{Evaluation for unseen category.}} For certain applications (\eg, tracking rare object classes with very few videos for training), it is desired to evaluate the performance of a tracker in locating targets belonging to previously unseen category. Current large-scale benchmark (\eg~\cite{muller2018trackingnet}) often have category overlaps between training and testing sequences, which makes it hard to meet this evaluation requirement. In order to alleviate this problem, the recently proposed GOT-10k~\cite{huang2018got} that takes the first attempt to introduce one-shot evaluation for tracking, aiming to assess tracking performance for unseen object classes.

In the literature, many benchmarks have been introduced to handle the aforementioned problems: \eg,~\cite{wu2013online,liang2015encoding,kristan2016novel,wu2015object,kristan2018visual,li2016nus} for precise dense annotations, \cite{mueller2016benchmark,valmadre2018long} for long-term tracking, \cite{huang2018got} for diverse object categories and unseen classes, and~\cite{muller2018trackingnet,huang2018got} for large-scale tracking. However, none of them address all of the issues, which motivates our proposed benchmark.

\subsection{Contribution}

In this work, we provide a novel benchmark for {\bf La}rge-scale {\bf S}ingle {\bf O}bject {\bf T}racking (LaSOT). The contributions of LaSOT are summarized as follows:

\renewcommand\arraystretch{1.05}
\begin{table*}[!t]\scriptsize
	\centering
	\caption{Comparison of LaSOT with the most popular dense benchmarks in the literature. ``Eva.'' and ``Tra.'' indicate evaluation and training, respectively. $^{*}$Note that, GOT-10k~\cite{huang2018got} covers more specific object and motion classes. For example, `person in jogging' and `person in skiing' are treated as two different object classes. However, in this work (and all other benchmarks), both belong to the category of `person'.}
	\begin{tabular}{rC{1.15cm}@{}C{1.25cm}@{}C{1.15cm}@{}C{1.15cm}@{}C{1.5cm}@{}C{1.5cm}@{}C{1.2cm}@{}C{1.35cm}@{}C{0.95cm}@{}C{1.5cm}@{}c}
		\hline
		Benchmark & \tabincell{c}{OTB-13 \\ \cite{wu2013online}} & \tabincell{c}{OTB-15 \\ \cite{wu2015object}} & \tabincell{c}{TC-128 \\ \cite{liang2015encoding}} & \tabincell{c}{CDTB \\ \cite{lukezic2019cdtb}}  & \tabincell{c}{VOT-2017 \\ \cite{kristan2017visual}} & \tabincell{c}{NUS-PRO \\ \cite{li2016nus}} & \tabincell{c}{UAV123 \\ \cite{mueller2016benchmark}} & \tabincell{c}{UAV20L \\ \cite{mueller2016benchmark}} & \tabincell{c}{NfS \\ \cite{galoogahi2017need}} & \tabincell{c}{GOT-10k \\ \cite{huang2018got}}  & LaSOT \\
		\hline
		\hline
		Num. of videos & 51    & 100   & 128   & 80    & 60    & 365   & 123   & 20    & 100   & 9,695 & 1,550 \\
		Min frames & 71    & 71    & 71    & 406     & 41    & 146   & 109   & 1,717 & 169   & 29 & 1,000 \\
		Mean frames & 578   & 590   & 429   & 1,274 & 356   & 371   & 915   & 2,934 & 3,830 & 149 &  2,502 \\
		Median frames & 392   & 393   & 365   & 1,179     & 293   & 300   & 882   & 2,626 & 2,448 & 101 & 2,145 \\
		Max frames & 3,872 & 3,872 & 3,872 & 2,501     & 1,500 & 5,040 & 3,085 & 5,527 & 20,665 & 1,418 & 11,397 \\
		Total frames & 29$\mathbf{K}$   & 59$\mathbf{K}$   & 55$\mathbf{K}$   & 102$\mathbf{K}$  & 21$\mathbf{K}$   & 135$\mathbf{K}$  & 113$\mathbf{K}$  & 59$\mathbf{K}$   & 383$\mathbf{K}$  & 1.45$\mathbf{M}$ &  3.87$\mathbf{M}$ \\
		Total duration & 16.4 $\mathbf{m}$ & 32.8 $\mathbf{m}$ & 30.7 $\mathbf{m}$ & 56.7 $\mathbf{m}$ & 11.9 $\mathbf{m}$ & 75.2 $\mathbf{m}$ & 62.5 $\mathbf{m}$ & 32.6 $\mathbf{m}$ & 26.6 $\mathbf{m}$ & 40 $\mathbf{h}$ & 35.8 $\mathbf{h}$ \\
		Video framerate & 30 fps    & 30 fps   & 30 fps   & 30 fps    & 30 fps    & 30 fps    & 30 fps    & 30 fps    & 240 fps   & 10 fps & 30 fps \\
		Object classes & 10     & 16    & 27    & 23     & 24    & 8     & 9     & 5     & 17     & 563$^{*}$ & 85 \\
		Num. of attributes & 11    & 11    & 11    & 13    & n/a     & n/a     & 12    & 12    & 9   &6   & 14 \\
		Absent labels &   \xmark    &   \xmark    &   \xmark    &   \cmark    &    \xmark   &   \xmark    &   \xmark    &   \xmark  & \xmark  &   \cmark    & \cmark \\
		Fully class balanced &  \xmark     &   \xmark    &   \xmark    &   \xmark    &    \xmark   &  \xmark     &     \xmark  &   \xmark    &    \xmark & \xmark  &  \cmark \\
		Axis alignment &  \cmark     &   \cmark    &   \cmark    &  \cmark    &    \xmark   &  \cmark     &     \cmark  &   \cmark   &    \cmark & \cmark  &  \cmark \\
		Lingual specification & \xmark    & \xmark    & \xmark    & \xmark    & \xmark     & \xmark     & \xmark    & \xmark    & \xmark   & \xmark   & \cmark \\
		Full overlap protocol & n/a    & n/a    & n/a    & n/a    & n/a     & n/a     & n/a    & n/a    & n/a   & \xmark   & \cmark \\
		One-shot protocol & n/a    & n/a    & n/a    & n/a    & n/a     & n/a     & n/a    & n/a    & n/a   & \cmark   & \cmark \\
		Benchmark aim & Eva.    & Eva.    & Eva.    & Eva.    & Eva.     & Eva.     & Eva.    & Eva.    & Eva.   & Tra./Eva.   & Tra./Eva. \\
		\hline
	\end{tabular}%
	\label{tab:DMA_comparison}%
\end{table*}%

%\vspace{-.6em}
\begin{enumerate}[(1)]
	\setlength{\itemsep}{1pt}
	\setlength{\parsep}{1pt}
	\setlength{\parskip}{1pt}
	
	\item We present a large-scale benchmark, LaSOT, for visual tracking. LaSOT covers 85 object categories and consists of 1,550 videos totaling more than 3.87M frames. Each frame is carefully inspected and manually labeled with a bounding box. To ensure quality, each annotation box is visually double-checked and corrected when needed. To our knowledge, LaSOT is by far the largest (in terms of the number of frames) tracking benchmark with precise dense annotations. By releasing LaSOT, we expect to offer the community a dedicated platform for unified training and evaluation of tracking algorithms.
	
	\item LaSOT allows evaluation of long-term tracking. In particular, the shortest sequence consists of 1,000 frames and the longest 11,397 frames, and the average video length of LaSOT is around 2,500 frames (equating to around 83 seconds, see Table~\ref{tab:DMA_comparison}), enabling assessment of long-term trackers. 
	
	\item Different from current benchmarks which only provide bounding boxes, LaSOT offers both visual bounding box annotations and natural language specification, which has been shown to be beneficial for various vision tasks (\eg, \cite{hu2016natural,li2017person}) including tracking~\cite{li2017tracking,feng2020real}. By providing additional language annotations, we aim at stimulating the use of lingual features to further improve tracking.
	
	\item For flexible evaluation of trackers in different settings, we adopt two protocols, \ie, \emph{full overlap} and \emph{one-shot}. For full overlap protocol, training and testing sets have the same object classes.  For one-shot protocol, as introduced in~\cite{huang2018got}, the categories of training and testing sets instead have zero overlap. These two protocols enable researchers/engineers to more flexibly evaluate their trackers to differing requirements, \eg, locating targets belonging to seen/unseen categories.
	
	\item LaSOT inhibits category bias by collecting equal numbers of videos for each object class\footnote{Note that for tracking benchmark using \emph{full overlap} split protocol, category bias should be inhibited in both training and evaluation of trackers. For tracking benchmark using \emph{one-shot} split protocol, category bias should be inhibited in only training of trackers.}. By doing this, the evaluation and comparison of trackers becomes more fair. To our knowledge, LaSOT is the first benchmark rigorously balanced for equal category size.
	
	\item To evaluate existing trackers and enable future comparison on LaSOT, we benchmark 48 representative tracking algorithms under the two protocols, and conduct extensive and in-depth analysis on performance using different metrics.
\end{enumerate}

This paper extends an early conference version in~\cite{fan2019lasot}. The main new contributions are follows. (\romannumeral1) We introduce 15 extra new object classes with 150 manually annotated sequences and more than 350K frames. In particular, different from classes chosen based on ImageNet in~\cite{fan2019lasot}, the 15 new classes are intentionally and carefully selected outside of ImageNet. By doing so, our benchmark enables a new \emph{one-shot} evaluation protocol using these 15 classes for testing. (\romannumeral2) More details of benchmark construction are provided. (\romannumeral3) We employ two different protocols, \emph{full overlap} and \emph{one-shot}, for flexible performance evaluation for seen/unseen target categories. (\romannumeral4) More thorough experimental analysis are conducted in various aspects.

The rest of this paper is organized as follows. Section 2 discusses related tracking algorithms and datasets of this work. In section 3, we detail the construction of LaSOT and analyze it through a variety of informative statistics. Experimental evaluation with in-depth analysis are conducted in section 4, followed by conclusion in section 5.

\section{Related Work}

\subsection{Visual Tracking Algorithm}

Visual tracking has been extensively studied in the past few decades. Here we briefly review two recent trends including correlation-filter trackers and deep trackers, and refer readers to surveys~\cite{li2013survey,smeulders2014visual,yilmaz2006object,li2018deep} for more algorithms.

Correlation-filter approaches formulate tracking task as a regression problem by learning a discriminative filter. Owing to the extremely efficient solution using fast Fourier transform (FFT), correlation-filter trackers~\cite{bolme2010visual,henriques2015high} run at speeds of several hundred frames per second and draw extensive attention with many improvements. The methods of~\cite{li2014scale,danelljan2014accurate} introduce a scale embedding to handle scale variation. The approaches in~\cite{danelljan2015learning,li2018learning} improve correlation-filter tracking using extra regularization techniques. Background information is explored in~\cite{mueller2017context,galoogahi2017learning} to enhance robustness of the filters. The methods of~\cite{ma2015hierarchical,danelljan2016beyond,danelljan2017eco} replace hand-crafted features with deep features to improve performance. The approach in~\cite{liu2015real} utilizes part-based representation to deal with challenges that are difficult for correlation filter tracking.

Inspired by the success of deep learning, many deep trackers~\cite{wang2013learning,wang2015visual,nam2016learning,fan2017sanet,song2018vital} have been proposed and exhibit state-of-the-art performance. Despite impressive results, these approaches suffer from heavy computational burden due to deep feature extraction or online network fine-tuning. To alleviate this problem, deep Siamese networks have been introduced for object tracking~\cite{bertinetto2016fully,tao2016siamese}. %Since model update is not required, Siamese trackers run efficiently in real-time. 
Owing to balanced efficiency and accuracy, deep Siamese tracking has been extended by many later works~\cite{he2018twofold,li2018high,fan2019siamese,wang2019spm,wang2019unsupervised,zhu2018distractor,li2019gradnet}. To deal with scale variation, the methods of~\cite{danelljan2019atom} introduce the intersection-over-union (IoU) network for tracking and achieve promising results.          

\subsection{Visual Tracking Benchmark}

Benchmarks have been crucial for advancing the research in visual tracking. For a systematic review, we  classify existing benchmarks into two types: \emph{dense benchmarks} which use per-frame manual annotation and \emph{other benchmarks} which use sparse and/or (semi-)automatic annotation.     

\subsubsection{Dense benchmarks}

Dense benchmarks offer \emph{per-frame} bounding box annotations for each video. In order to ensure high quality, each frame is manually annotated with careful inspection and verification. For tracking, these precise bounding box annotations are highly desired for both training and evaluating tracking algorithms. Currently, popular dense tracking benchmarks include OTB~\cite{wu2013online,wu2015object}, TC-128~\cite{liang2015encoding}, VOT~\cite{kristan2016novel}, NUS-PRO~\cite{li2016nus}, UAV~\cite{mueller2016benchmark},  NfS~\cite{galoogahi2017need}, CDTB~\cite{lukezic2019cdtb} and GOT-10k~\cite{huang2018got}.

\vspace{0.2em}
\noindent {\bf OTB.} OTB-13~\cite{wu2013online} contains 51 videos with manual annotation for tracking evaluation. The videos are labeled with 11 attributes for further analysis of tracking performance. OTB-13 was later extended to the larger OTB-15~\cite{wu2015object} by introducing extra 50 sequences.

\vspace{0.2em}
\noindent {\bf TC-128.} TC-128~\cite{liang2015encoding} comprises of 128 videos that are specifically designated to evaluate color-enhanced trackers. The videos are labeled with 11 similar attributes as in OTB~\cite{wu2013online}.

\vspace{0.2em}
\noindent {\bf VOT.} VOT~\cite{kristan2016novel} introduces a series of tracking competitions with up to 60 sequences in each of them, aiming to evaluate the performance of a tracker in a relative short duration. Each frame in the VOT datasets is annotated with a rotated bounding box with several attributes.

\vspace{0.2em}
\noindent {\bf CDTB.} CDTB~\cite{lukezic2019cdtb} offers 80 RGB-D videos with manual annotations for tracking. Each sequence is labeled with 13 attributes. The goal of CDTB is to encourage the exploration of depth information for improving tracking performance.

\vspace{0.2em}
\noindent {\bf NUS-PRO.} NUS-PRO~\cite{li2016nus} contains 365 sequences with a focus on human and rigid object tracking.  Each sequence in NUS-PRO is annotated with both target location and occlusion level for evaluation.

\vspace{0.2em}
\noindent {\bf UAV.}  UAV123 and UAV20L~\cite{mueller2016benchmark} are utilized for unmanned aerial vehicle (UAV) tracking, comprising 123 short and 20 long sequences, respectively. Both UAV123 and UAV20L are labeled with 12 attributes.

\vspace{0.2em}
\noindent {\bf NfS.} NfS~\cite{galoogahi2017need} provides 100 sequences with a high frame rate of 240 fps, aiming to analyze the effects of appearance variations on tracking performance.

\vspace{0.2em}
\noindent {\bf GOT-10k.} GOT-10k~\cite{huang2018got} consists of 9,695 videos, aiming to provide rich motion trajectories for developing and evaluating trackers. In addition, GOT-10k is the first to propose a novel one-shot evaluation for assessing tracking performance.

Our LaSOT belongs to the category of dense tracking benchmark. In comparison with others, LaSOT is the \emph{largest} with more than 3.87 million frames and an average video length of around 2,500 frames. Moreover, LaSOT is the only one to offer additional language specification for each sequence. LaSOT is closely related to but different from the recently proposed large-scale GOT-10k~\cite{huang2018got}. Despite sharing the similar idea of performing one-shot evaluation, LaSOT presents two protocols. In addition, instead of focusing on short-term tracking GOT-10k, our goal is to assess trackers in long-term scenarios. Table~\ref{tab:DMA_comparison} provides a detailed comparison of LaSOT with existing dense benchmarks.  It is worth noting that most existing dense tracking benchmarks, including LaSOT, utilize axis-aligned bounding boxes to annotate targets. The reasons are two-fold. First, the problem setting of current single object tracking is to locate the target with a manually given up-right bounding box. In accordance with this goal, axis-aligned boxes are usually adopted to annotate targets in many benchmarks. Axis-aligned boxes are also widely employed in object detection benchmarks such as PASCAL VOC~\cite{everingham2010pascal} and COCO~\cite{lin2014microsoft}. Second, axis-aligned boxes are able to provide sufficient information about the target for stable tracker initialization and reliable performance evaluation, as evidenced by recent progresses of tracking algorithms on various benchmarks. From this perspective, axis-aligned boxes are effective for tracking. Moreover, this type of annotation requires less labeling efforts.

\subsubsection{Other tracking benchmarks}

Aside from benchmarks described above, there are other benchmarks using different annotation strategies. These tracking benchmarks are either labeled sparsely (\eg, every 30 frames) or annotated (semi)-automatically using tracking algorithms. Examples of these types of benchmarks include ALOV~\cite{smeulders2014visual}, TrackingNet~\cite{muller2018trackingnet} and OxUvA~\cite{valmadre2018long}.

{\bf ALOV}~\cite{smeulders2014visual} comprises of 314 video sequences which are labeled in 14 attributes. Instead of per-frame annotation, ALOV provides annotations every 5 frames. {\bf TrackingNet}~\cite{muller2018trackingnet} is a large-scale benchmark with 30K sequences. All videos come from the video object detection dataset YT-BB~\cite{real2017youtube}, and each one is labeled by a tracking algorithm. Although this tracker annotator is shown to be reliable in a relatively short period (\ie, 1 second), it is hard to guarantee the same tracking performance on a different benchmark, especially when the sequences become more challenging. In addition, the average video length of TrackingNet is less than 500 frames, which may not be able to reflect the long-term performance of a tracking algorithm. {\bf OxUvA}~\cite{valmadre2018long} consists of 366 sequences. Similar to TrackingNet, the videos are sampled from YT-BB~\cite{real2017youtube}. With the average sequence length more than 4,200 frames, OxUvA mainly aims to focus on long-term tracking. Each video in OxUvA is labeled every 30 frames.

% With a goal of facilitating long-term tracking, its average sequence length exceeds 4,200 frames. An issue with OxUvA, nevertheless, is that it does not densely annotate each frame in a video. Each sequence is labeled every 30 frames, ignoring rich motion information and temporal context in videos when developing tracking approaches.

These benchmarks usually provide a large number of sequences and serve well for evaluation purposes. While they benefit from a reduction of annotation cost, they do not provide detailed per frame performance evaluation of tracking algorithms. Furthermore, it may cause problems for some trackers that require temporal context or motion cues from annotations, because these information may be either missing due to sparse annotation or imprecise due to potentially unreliable annotation. Different from these benchmarks, LaSOT provides a large set of sequences with high-quality dense bounding box annotations, which makes it more suitable for developing deep trackers as well as for evaluating long-term tracking algorithms.

\subsection{Other Vision Benchmarks}

Given the similarities shared between visual object tracking and video object detection (\eg, visual tracking can be treated as video single-object detection), video object detection benchmarks VID~\cite{russakovsky2015imagenet} and YT-BB~\cite{real2017youtube} are often adopted for training deep trackers.

{\bf VID}~\cite{russakovsky2015imagenet} consists of 5.4K sequences with more than two million frames and {\bf YT-BB}~\cite{real2017youtube} contains 380K videos with more than five million frames. Despite being large in scale, these two benchmarks are not ideally suitable for tracking tasks due to several reasons. First, in many videos, the targets are almost static throughout the entire video, making them not desirable for motion tracking. Second, the targets are partially out of view in the initial frame in a lot of videos, which is different from the tracking task. Third, the benchmarks are sparsely annotated, and thus may be inappropriate if directly used for tracking as discussed early. 

%\subsection{{Other Vision Benchmarks}}

In the era of deep learning, benchmarks have played a more important role in advancing various vision tasks. To some extent, LaSOT is inspired by the successes of other vision benchmarks. To this end, we will briefly discuss several large-scale benchmarks in other tasks including image classification, object detection, segmentation and multi-object tracking.

In image classification, {\bf ImageNet}~\cite{deng2009imagenet} is  arguably the most popular dataset consisting of more than 10M images. Owing to the large-scale ImageNet, deep networks have proven their power in learning visual representation. In object detection, the well-known {\bf PASCAL VOC} detection~\cite{everingham2010pascal} contains around 10K images. The larger scale {\bf COCO}~\cite{lin2014microsoft} contains more than 200K images for detection. In image segmentation, {\bf PASCAL VOC} segmentation~\cite{everingham2010pascal} provides around 10K images. {\bf ADE20K}~\cite{zhou2017scene} is a collection of more than 20K images for scene parsing. {\bf Citiscapes}~\cite{cordts2016cityscapes} consists of 25K images for traffic scene segmentation. {\bf LVIS}~\cite{gupta2019lvis} offers 164K image for large-scale vocabulary instance segmentation. In multi-object tracking, the {\bf MOT} challenge~\cite{milan2016mot16} provides 21 videos. Recently, a larger scale {\bf TAO}~\cite{dave2020tao} has been compiled containing 2,907 videos.

\section{The LaSOT Benchmark}

%In this section, we detail the construction of LaSOT with a variety of informative statistics.

\subsection{Design Principle}

Our goal is to construct a dedicated benchmark, LaSOT, for training and evaluating tracking algorithms. To this end, we follow six principles in constructing LaSOT, including \emph{large-scale}, \emph{high-quality dense annotations}, \emph{long-term tracking}, \emph{category balance}, \emph{comprehensive labeling} and \emph{flexible protocols}, aimed at handling the issues of existing tracking benchmarks described in previous sections.

\begin{figure*}[!t]
	\centering
	\includegraphics[width=0.99\linewidth]{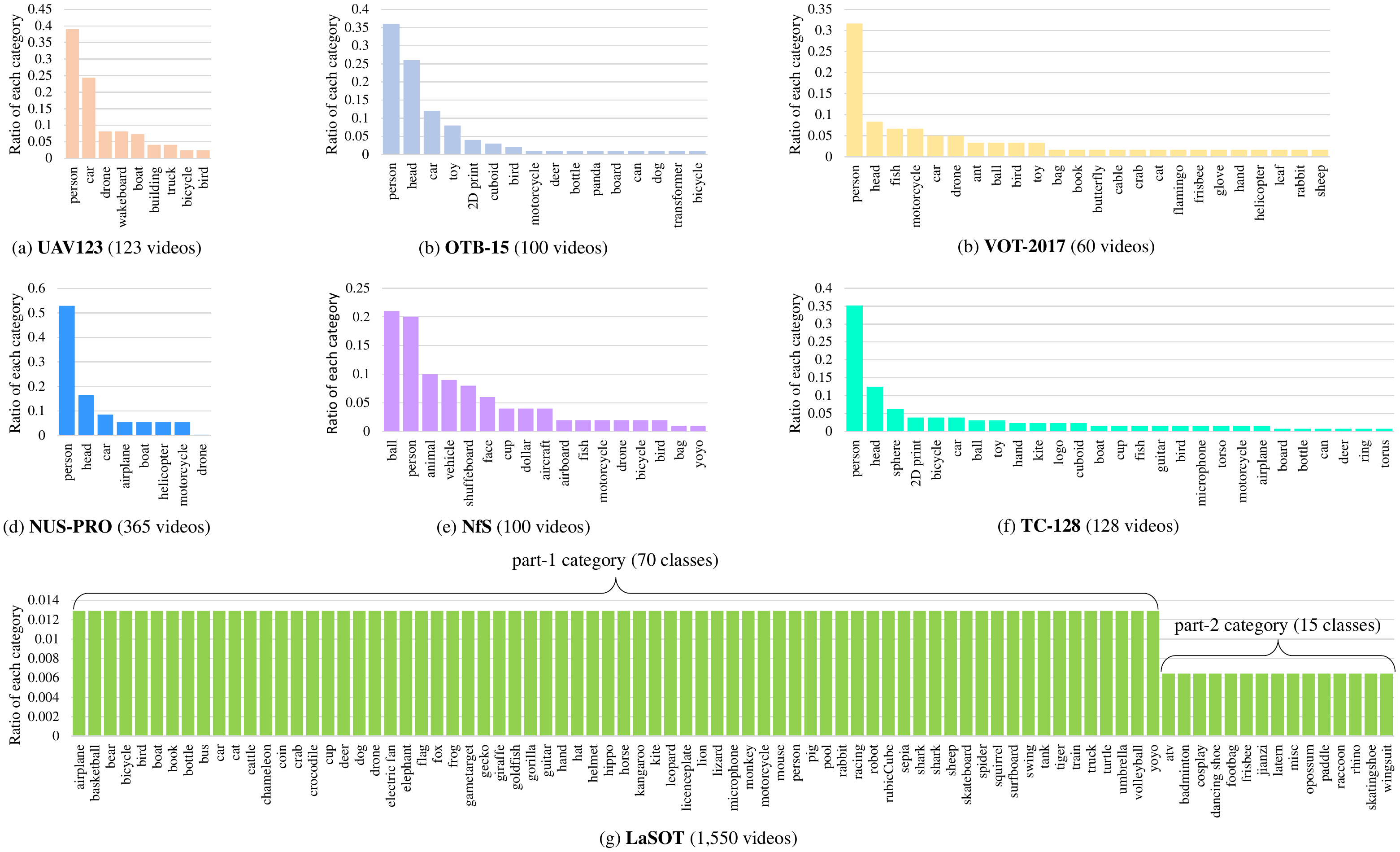}\\
	\caption{Category distribution of tracking benchmarks. The category distribution of LaSOT is more balanced than those of other benchmarks. Best viewed in color.}
	\label{fig:benchmark-category}
\end{figure*}

\subsection{Data Collection}

In total, LaSOT consists of 85 object classes, which are divided into two parts. The first part, referred to as \emph{part-1} for short, contains 1,400 sequences from 70 object categories. Most of categories are chosen from the 1,000 classes from ImageNet~\cite{deng2009imagenet}, with a few exceptions (\eg, \emph{drone}) that are carefully selected for popular tracking applications. The other part, referred to as \emph{part-2} for short, comprises 150 sequences from 15 object classes. It is worth noting that, for the goal of one-shot evaluation on object from unseen categories, these 15 classes are carefully chosen from \textbf{\emph{outside}} object categories in ImageNet~\cite{deng2009imagenet} and intentionally to be far away from the 70 categories in part-1. There is no overlap between the 15 categories in part-2 and 70 classes in part-1. Different from current dense tracking benchmarks that contain less than 30 categories and typically are unevenly distributed, LaSOT provides equal number of videos for each category in both part-1 and part-2 to avoid the category bias problem. 

After determining the 85 object classes in LaSOT, we searched for sequences of each category from YouTube (\url{https://www.youtube.com/}). The reasons for choosing YouTube are two-fold: (1) YouTube is the largest video platform in the world, which allows us to select diverse videos for constructing the benchmark and avoids bias to certain scenes, and (2) many videos on YouTube are captured in the wild, which may be helpful for developing and evaluating trackers for real applications.

Initially, over 6,000 video sequences are collected. With a joint consideration of the video quality (\eg, videos with shot cut are not suitable for tracking) and our design principles, 1,550 sequences survived. Nevertheless, these 1,550 videos are not immediately available for the tracking task due to containing a large mount of irrelevant contents. For instance, for a video of \emph{person} category (\eg, a sporter), it often consists of some undesirable introduction content of each sporter in the beginning. Therefore, we carefully inspect each video sequence, filter out the tracking-unrelated contents and exclusively retain one usable clip for our tracking task. For  part-1, each category consists of 20 videos, while for part-2, each contains 10 sequences. Figure~\ref{fig:benchmark-category} shows the object categories on LaSOT with comparison to several existing popular dense tracking benchmarks with available category information. It is worth noting that, although the numbers of videos for categories in part-1 and part-2 are not equal, LaSOT is still balanced due to their different roles as described in section 3.5. Also note that, in Figure~\ref{fig:benchmark-category} we do not include the large-scale GOT-10k for comparison because the category granularity used in GOT-10k is different from those in other benchmarks. For example, ``big truck '', ``half truck'' and ``pickup truck'' are treated as three different categories in GOT-10k. By contrast, in other benchmarks, there may exist only one ``truck'' category. %To this end, we do not compare with GOT-10k on category distribution.

Eventually, we compiled a large-scale tracking benchmark by gathering 1,550 videos with 3.87 million frames from YouTube under Creative Commons license. The average video length of LaSOT is 2,502 frames (\ie, 83 seconds for 30 fps). The shortest sequence contains 1,000 frames (\ie, 33 seconds), while the longest one consists of 11,397 frames (\ie, 378 seconds).

\subsection{Annotation}

%In the construction of LaSOT, the annotation part requires the most effort. In this section, we will detail the annotation and quality assessment protocols of LaSOT.

\subsubsection{Annotation protocol}

\begin{figure}[!tb]
	\centering
	\includegraphics[width=0.98\linewidth]{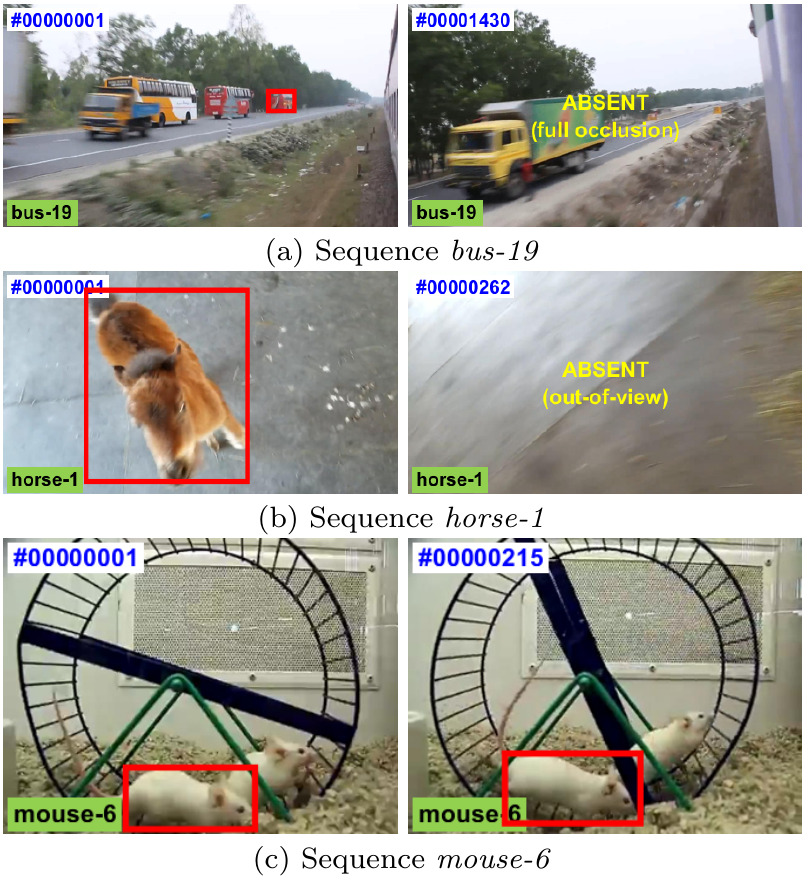}\\
	\caption{Illustration of annotation strategy for different cases. Image (a): target with full occlusion. Image (b): target with out-of-view. Image (c): target with thin, long, and highly deformable part. Best viewed in color.}
	\label{fig:three_annotation_label}
\end{figure}

Annotation consistency cross different sequences and labelers is crucial for the quality of a tracking benchmark. We define a deterministic protocol for ensuring such quality. In a video sequence with a specific tracking target (determined before starting annotation), for each frame, if the target is present in the view, a labeler manually draws/edits an up-right (axis-aligned) bounding box to tightly fit any visible part of the target (see left images of (a) and (b) in Figure~\ref{fig:three_annotation_label}); otherwise, an absence label, either \emph{full occlusion} (see right image of (a) in Figure~\ref{fig:three_annotation_label}) or \emph{out-of-view} (see right image of (b) in Figure~\ref{fig:three_annotation_label}), is assigned to this frame. By doing so, there are two advantages: (1) with absence labels, performance evaluation is more accurate by avoiding those frames without target present, and (2) researchers can develop occlusion or out-of-view aware tracking algorithms using this information. Note that, our strategy cannot guarantee to minimize the background area in the box, as similarly observed in other benchmarks. Nevertheless, this strategy provides consistent annotations that are relatively stable for learning the dynamics.

The above annotation strategy works well most of the time, however, exceptions exist. For certain categories, \eg, \emph{mouse}, the target object may contain long, thin, and/or highly deformable parts, \eg, a tail, which not only introduces much background information into object, but also provides little help for target recognition and localization. We carefully identify such targets and associated videos in LaSOT, and design specific rules for their annotations. In detail, before starting to annotate, we inspect each object category and identify twelve such categories and their undesired parts, including {\it bird} (the {\it leg} part), {\it cat} (the {\it tail} part), {\it elephant} (the {\it tail} part), {\it fox} (the {\it tail} part), {\it gecko} (the {\it tail} part), {\it guitar} (the {\it handlebar} part), {\it leopard} (the {\it tail} part), {\it lion} (the {\it tail} part), {\it monkey} (the {\it tail} part), {\it tiger} (the {\it tail} part), {\it shark} (the {\it tail} part) and {\it mouse} (the {\it tail} part). For objects from these categories, we exclude the undesired part when drawing their bounding boxes. Note that, to ensure the usability of these classes, the inspection of each object category and identification of undesired parts are conducted by a group of experts (three PhD students working in related areas). An annotation example is shown in the image (c) of Figure~\ref{fig:three_annotation_label} and more can be found in our benchmark.

In order to enrich annotation, we provide additional language descriptions for each sequence. The natural language specification is represented by a sentence that describes the color, behavior and surroundings of the target. LaSOT consists of 1,550 such sentences for all sequences. Notice that, we expect that these lingual descriptions can provide auxiliary help for improving tracking. For example, one can leverage deep neural networks to extract lingual features and use them as a global semantic guidance to suppress background distractors in the search region. This way, the tracker may better focus on locating the target.

\subsubsection{Quality assessment protocol}

For developing a high-quality dense benchmark, the most effort demanding parts include \emph{manual labeling}, \emph{double-checking} and \emph{error correcting}. For this task, we have assembled an annotation team composed of several Ph.D. students working on related areas and many volunteers. To ensure high-quality annotation, each video is processed by two teams: a labeling team and a validation team. Each labeling team is composed of a volunteer and an expert (Ph.D. student). The volunteer manually draws/edits the target bounding box in each frame, and the expert inspects the results and adjusts them if necessary. Then, the annotation results are reviewed by the validation team composed of several (typically three) experts. If an annotation result is not unanimously agreed by all members in the validation team, it will be sent back to the original labeling team to revise. Note that, when sending the annotation results back for revision, detailed comments from the validation team are attached. Examples include ``the annotated bounding box is too small to cover the whole target,"  ``the box is too large and introduce too much background," ``the left  side contains much background and its edge needs to move closer the target boundary," etc. This way, we ensure that the revised annotation result is acceptable as expected.

\begin{figure}
	\centering
	\includegraphics[width=0.99\linewidth]{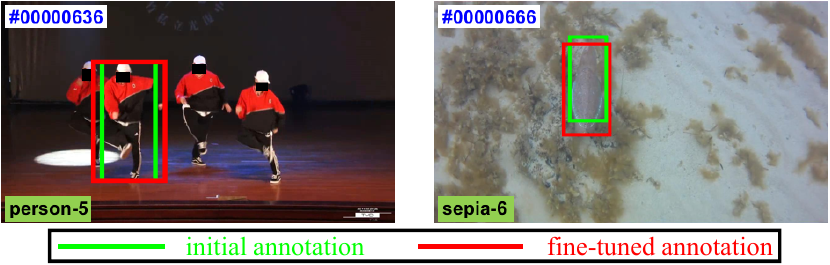}\\
	\caption{Examples of fine-tuning initial annotations. We observe that, after fine-tuning, the final annotations in red rectangles better fit to the target region than the initial annotations in green rectangle. Best viewed in color.
	}
	\label{annotation-revision}
\end{figure}

\begin{figure*}[!t]
	\centering
	\includegraphics[width=\linewidth]{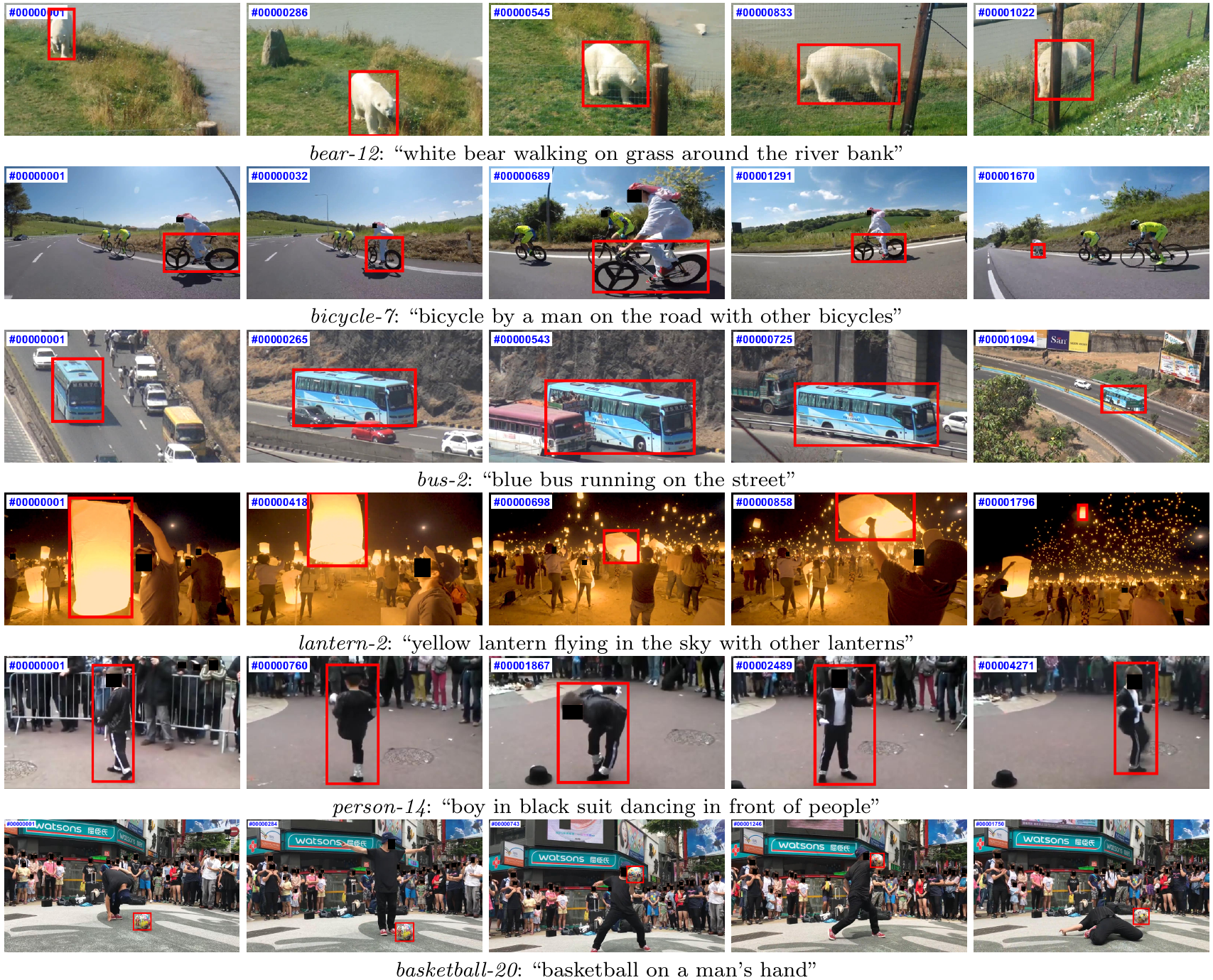}\\
	\caption{Example sequences and annotations in LaSOT. Best viewed in color.}
	\label{annotation_sample}
\end{figure*}

\renewcommand\arraystretch{1.05}
\begin{table*}[!t]
	\centering
	\caption{Descriptions of 14 different attributes in LaSOT.}
	\begin{tabular}{rl}
		\hline
		Attribute & Definition  \\
		\hline \hline
		{\bf CM }   & \underline{\emph{Camera Motion}}: abrupt motion of the camera  \\
		{\bf VC }   & \underline{\emph{View Change}}: viewpoint affects target appearance significantly \\
		{\bf ROT }  & \underline{\emph{Rotation}}: the target object rotates in the image \\ 
		{\bf SV }   & \underline{\emph{Scale Variation}}: the ratio of target bounding box is outside the range [0.5, 2] \\
		{\bf DEF }  & \underline{\emph{Deformation}}: the target object is deformable during tracking \\ 
		{\bf BC }   & \underline{\emph{Background Clutter}}: the background near the target object has the similar appearance as the target \\
		{\bf POC }  & \underline{\emph{Partial Occlusion}}: the target object is partially occluded in the sequence \\ 
		{\bf FOC }  & \underline{\emph{Full Occlusion}}: the target object is fully occluded in the sequence \\ 
		{\bf MB }   & \underline{\emph{Motion Blur}}: the target region is blurred due to the motion of target object or camera \\
		{\bf IV }   & \underline{\emph{Illumination Variation}}: the illumination in the target region changes \\
		{\bf ARC }  & \underline{\emph{Aspect Ratio Change}}: the ratio of bounding box aspect ratio is outside the rage [0.5, 2] \\
		{\bf OV }   & \underline{\emph{Out-of-View}}: the target object completely leaves the video frame \\ 
		{\bf LR }   & \underline{\emph{Low Resolution}}: the area of target box is smaller than 1000 pixels in at least one frame \\
		{\bf FM }   & \underline{\emph{Fast Motion}}: the motion of target object is larger than the size of its bounding box \\
		\hline
	\end{tabular}%
	\label{tab:att_def}%
\end{table*}%

\begin{figure*}[!t]
	\centering
	\includegraphics[width=\linewidth]{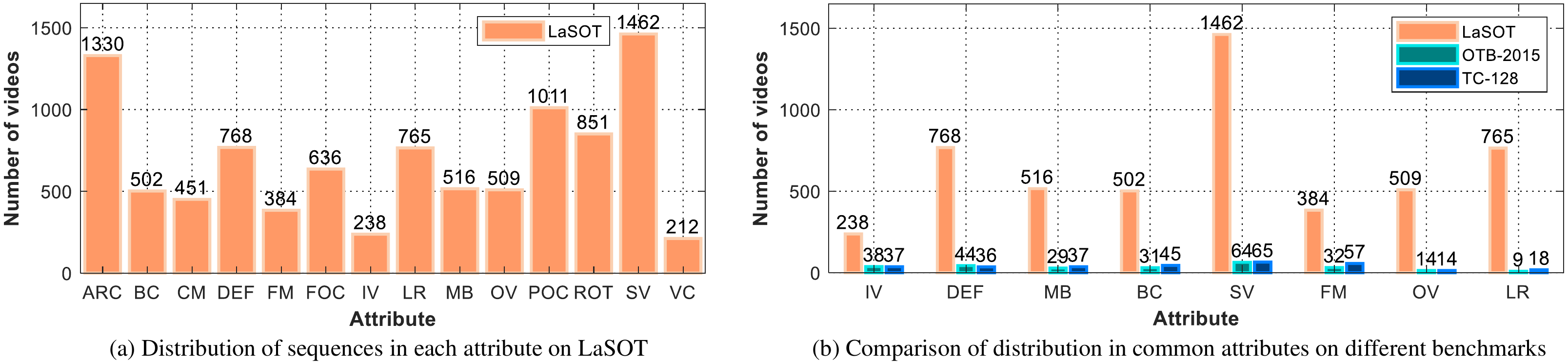}\\
	\caption{Distribution of sequences in each attribute in LaSOT and comparison with other benchmarks. Best viewed in color.}
	\label{fig:att_dis}
\end{figure*}

\renewcommand\arraystretch{1.05}
\begin{table}[htbp]
	\centering
	\caption{Comparisons between training/testing sets of LaSOT under {\bf full overlap} protocol.}
	\begin{tabular}{@{}rccccc@{}}
		\hline
		& Video &\tabincell{c}{Min \\ frames} & \tabincell{c}{Mean \\ frames}  & \tabincell{c}{Max \\ frames} & \tabincell{c}{Total \\ frames}  \\
		\hline \hline
		LaSOT$_{\mathrm{tra}}$ & 1,120  & 1,000  & 2,529 &  11,397 & 2.83$\mathbf{M}$   \\
		LaSOT$_{\mathrm{tst}}$ & 280   & 1,000  & 2,448 &  9,999 & 690$\mathbf{K}$   \\
		\hline
	\end{tabular}%
	\label{tab:full_overlap_training_testing}%
\end{table}%

To improve the annotation quality as much as possible, we check all the annotation results carefully and revise them frequently. Around 40\% of the initial annotations fail in the first round of validation, and many frames are revised at least three times. Some challenging frames that are initially labeled incorrectly or inaccurately are given in Figure~\ref{annotation-revision}. With all these efforts, we finally reach a benchmark with high-quality dense annotation, with some examples shown in Figure~\ref{annotation_sample}.

\subsection{Attributes}

In order to further analyze the tracking performance, each sequence in LaSOT is labeled with a list of 14 attributes, including camera motion (CM), rotation (ROT), deformation (DEF), full occlusion (FOC), partial occlusion (POC), illumination variation (IV), out-of-view (OV), viewpoint change (VC), scale variation (SV), background clutter (BC), motion blur (MB), aspect ratio change (ARC), low resolution (LR) and fast motion (FM). Table~\ref{tab:att_def} lists the definition of each attribute, and Figure~\ref{fig:att_dis} (a) shows the distribution of sequences in each attribute. 
From Figure~\ref{fig:att_dis} (a), it can be seen that the most common challenge factors in LaSOT are target scale changes (SV and ARC), occlusion (POC and FOC), deformation and rotation, which frequently occur in real applications.

In addition, Figure~\ref{fig:att_dis} (a) shows that each attribute consists of \emph{at least} 200 videos, which clearly supports the statistical significance of our attribute evaluation. Figure~\ref{fig:att_dis} (b) demonstrates the distribution of attributes of LaSOT compared with popular benchmarks OTB-15~\cite{wu2015object} and TC-128~\cite{liang2015encoding} on overlapping attributes. From Figure~\ref{fig:att_dis} (b), we observe that more than 1,400 videos in LaSOT are involved with scale variations. Compared with OTB-2015 and TC-128 with less than 70 videos with scale changes, LaSOT is more challenging and thus better reflects the generalizability of trackers in dealing with scale changes. On the out-of-view attribute, LaSOT contains 509 videos, while OTB-15 and TC-128 have less than 20 sequences, indicating that LaSOT reflects better the challenges for tracking in the wild. Moreover, LaSOT focuses on small object tracking with 765 videos in the attribute of low resolution, much more than that in OTB-15 and TC-128.

\renewcommand\arraystretch{1.05}
\begin{table}[!t]
	\centering
	\caption{Comparisons between training/testing sets of LaSOT under {\bf one-shot} protocol.}
	\begin{tabular}{@{}rccccc@{}}
		\hline
		& Video &\tabincell{c}{Min \\ frames} & \tabincell{c}{Mean \\ frames}  & \tabincell{c}{Max \\ frames} & \tabincell{c}{Total \\ frames}  \\
		\hline \hline
		LaSOT$_{\mathrm{tra}}$ & 1,400  & 1,000  & 2,506 &  11,397 & 3.52$\mathbf{M}$   \\
		LaSOT$_{\mathrm{tst}}$ & 150   & 2,005  & 2,393 &  2,500 & 350$\mathbf{K}$   \\
		\hline
	\end{tabular}%
	\label{tab:one_shot_training_testing}%
\end{table}%

It is worth noting that in our benchmark, as well as in most other popular ones, a video sequence may consist of more than one attribute. As a consequence, it may be difficult to concretely identify the attribute causing failure, especially if the number of videos on this attribute for evaluation is small. The ideal situation for attribute evaluation would be that each sequence exhibits {\it one and only one} attribute. Nevertheless, in real world applications, it is almost impossible for a video to contain only one challenge. To alleviate this problem and reduce uncertainty, in existing tracking benchmarks, a common way is to collect all videos containing a specific attribute when performing evaluation for that attribute. For example, one can usually gather more than thirty videos for some attributes. Especially, in our large-scale benchmark, the numbers of videos for most attributes exceed one hundred. Consequently, we may obtain a {\it statistically meaningful} conclusion for attribute evaluation despite that videos may contain mixed attributes. This is supported by the fact that many trackers with higher attribute evaluation scores generally work better in dealing with corresponding attributes in videos on various benchmarks. For this reason, following the studies in previous tracking benchmarks, attribute-based evaluation is conducted on LaSOT as well. With that said, it is worth noting a recent effort to restrict one attribute per (short) sequence in tracking evaluation~\cite{Fan20tracklinic}.

\subsection{Evaluation Protocols}

Currently, evaluation of large-scale benchmarks is based on either \emph{full overlap} (\eg,~\cite{muller2018trackingnet}) or \emph{one-shot} (\eg,~\cite{huang2018got}). We argue that both protocols have their own applications. The full overlap protocol splits training/testing sets with fully overlapped object classes, and it can be used to develop tracking algorithms in the scene where the target category appears in the tracker's training set. By contrast, one-shot protocol splits training/testing with no overlap between their object categories, and it can be utilized in applications where the target category is rare. In order to accommodate more application scenarios, we introduce both protocols into LaSOT. %, and researchers can adopt either one or both for evaluating their algorithms.

\renewcommand\arraystretch{1.1}
\begin{table*}[!t]
	\centering
	\caption{Summary of evaluated trackers. Representation: Sparse - Sparse Representation, Color - Color Names or Histograms, Pixel - Pixel Intensity, HoG - Histogram of Oriented Gradients, H or B - Haar or Binary, Deep - Deep Features, Update - Online model update (or fine-tuning). Search: PF - Particle Filter, RS - Random Sampling, DS - Dense Sampling.}
	\begin{tabular}{@{}rrccccccccccc@{}}
		\hline
		\multicolumn{1}{r}{\multirow{3}[0]{*}{ }} & \multicolumn{1}{r}{\multirow{3}[0]{*}{ }} & \multicolumn{8}{c}{\bf Representation}            & \multicolumn{3}{c}{\bf Search}  \\
		\cmidrule(lr){3-10}  \cmidrule(lr){11-13}
		&       & \rotatebox{0}{PCA} & \rotatebox{0}{Sparse} & \rotatebox{0}{Color} & \rotatebox{0}{Pixel} & \rotatebox{0}{HoG} & \rotatebox{0}{H or B} & \rotatebox{0}{Deep} & \rotatebox{0}{Update} & \rotatebox{0}{PF} & \rotatebox{0}{RS} & \rotatebox{0}{DS}   \\
		\hline \hline
		IVT~\cite{ross2008incremental}           & IJCV08 &    \cmark   &      &     &    &       &       & &   \cmark   &   \cmark    &        &      \\
		MIL~\cite{babenko2009visual}             & CVPR09 &             &       &       &   &     &   H   & &   \cmark   &       &       &   \cmark       \\
		Struck~\cite{HareST11}                   & ICCV11 &             &       &       &   &     &  H    &  &  \cmark   &       &       &   \cmark     \\
		L1APG~\cite{bao2012real}                 & CVPR12 &             &   \cmark    &    &    &       &   &    &  \cmark     &  \cmark   &       &         \\
		ASLA~\cite{jia2012visual}                & CVPR12 &             &   \cmark    &    &    &       &       &  &  \cmark   &   \cmark   &       &         \\
		CSK~\cite{henriques2012exploiting}       & ECCV12 &       &       &       & \cmark  &     &       &       & \cmark &     &       &   \cmark      \\
		CT~\cite{zhang2012real}                  & ECCV12 &       &       &       &   &     &   H    &    &  \cmark &      &       &   \cmark      \\
		TLD~\cite{kalal2012tracking}             & PAMI12 &       &       &       &  &      &  B     &   &  \cmark  &       &       &   \cmark      \\
		CN~\cite{danelljan2014adaptive}          & CVPR14 &       &       &   \cmark    & \cmark  &     &   &    &   \cmark    &       &       &    \cmark     \\
		DSST~\cite{danelljan2014accurate}        & BMVC14 &       &       &       & \cmark  &  \cmark   &   &    &   \cmark    &       &       &   \cmark      \\
		MEEM~\cite{zhang2014meem}                & ECCV14 &       &       &       & \cmark  &     &       &  &   \cmark  &       &  \cmark     &         \\
		STC~\cite{zhang2014fast}                 & ECCV14 &       &       &       & \cmark  &     &       &    &  \cmark &       &       &    \cmark     \\
		SAMF~\cite{li2014scale}                  & ECCVW14 &       &       &   \cmark    & \cmark  & \cmark    &    &   &   \cmark    &       &       &   \cmark      \\
		LCT~\cite{ma2015long}                    & CVPR15 &       &       &       &  \cmark  &  \cmark  &       &   &  \cmark  &       &       &   \cmark      \\
		SRDCF~\cite{danelljan2015learning}       & ICCV15 &       &       &       &   &  \cmark   &       &       &  \cmark &    &       &    \cmark     \\
		HCFT~\cite{ma2015hierarchical}           & ICCV15 &       &       &       &   &     &       &    VGG-19   & \cmark &     &       &    \cmark     \\
		KCF~\cite{henriques2015high}             & PAMI15 &       &       &       &  &  \cmark    &       &       & \cmark &     &       &   \cmark      \\
		Staple~\cite{bertinetto2016staple}       & CVPR16 &       &       &   \cmark    &   &  \cmark   &       &    & \cmark  &       &       &    \cmark     \\
		SINT~\cite{tao2016siamese}               & CVPR16 &       &       &       &   &     &       &   VGG-16    &  &     &    \cmark   &         \\
		SCT4~\cite{choi2016visual}               & CVPR16 &       &       &       &   &  \cmark   &       &       & \cmark  &    &       &   \cmark      \\
		MDNet~\cite{nam2016learning}             & CVPR16 &       &       &       &    &    &       &    VGG-M   & \cmark  &   &    \cmark    &       \\
		SiamFC~\cite{bertinetto2016fully}        & ECCVW16&       &       &       &   &     &       &    AlexNet    &   &    &       &     \cmark     \\
		Staple\_CA\cite{mueller2017context}      & CVPR17 &       &       &    \cmark   &    &  \cmark  &       &     & \cmark &       &       &    \cmark     \\
		ECO\_HC~\cite{danelljan2017eco}          & CVPR17 &       &       &       &   &   \cmark  &       &       &  \cmark &    &       &    \cmark     \\
		ECO~\cite{danelljan2017eco}              & CVPR17 &       &       &       &   &     &       &  VGG-M     &  \cmark &    &       &    \cmark     \\
		CFNet~\cite{valmadre2017end}             & CVPR17 &       &       &       &   &     &       &   AlexNet    & \cmark  &    &       &    \cmark     \\
		CSRDCF~\cite{lukezic2017discriminative}  & CVPR17 &       &       &  \cmark     & \cmark  &  \cmark   &     &  &  \cmark     &       &       &   \cmark      \\
		PTAV~\cite{fan2017parallel}              & ICCV17 &       &       &       & \cmark &  \cmark    &      & VGG-16  &  \cmark  &       &       &     \cmark    \\
		DSiam~\cite{guo2017learning}             & ICCV17 &       &       &       &    &    &       &   AlexNet   &  &     &       &   \cmark      \\
		BACF~\cite{galoogahi2017learning}        & ICCV17 &       &       &       &    & \cmark   &       &       & \cmark &     &       &   \cmark      \\
		fDSST~\cite{danelljan2017discriminative} & PAMI17 &       &       &       & \cmark  &  \cmark   &       &   & \cmark  &       &       &    \cmark     \\
		VITAL~\cite{song2018vital}               & CVPR18 &       &       &       &    &    &       &    VGG-M   &  \cmark &    &   \cmark         \\
		TRACA~\cite{choi2018context}             & CVPR18 &       &       &       &   &     &       &  VGG-M     & \cmark  &    &       &    \cmark     \\
		STRCF~\cite{li2018learning}              & CVPR18 &       &       &       &   &   \cmark  &       &       & \cmark  &    &       &    \cmark     \\
		D-STRCF~\cite{li2018learning}              & CVPR18 &       &       &       &   &     &       &  VGG-M     & \cmark  &    &       &    \cmark     \\
		StructSiam~\cite{Zhang2018structured}    & ECCV18 &       &       &       &    &    &       &   AlexNet    &   &    &       &   \cmark     \\
		DaSiamRPN~\cite{zhu2018distractor}    & ECCV18 &       &       &       &    &    &       &   Res-50    & \cmark  &    &       &   \cmark     \\
		SiamRPN++~\cite{li2019siamrpn++}    & CVPR19 &       &       &       &    &    &       &   Res-50    &   &    &       &   \cmark     \\
		SiamDW~\cite{zhang2019deeper}    & CVPR19 &       &       &       &    &    &       &   Res-22    &   &    &       &   \cmark     \\
		SiamMask~\cite{wang2019fast}    & CVPR19 &       &       &       &    &    &       &   Res-50    &   &    &       &   \cmark     \\
		ASRCF~\cite{dai2019visual}    & CVPR19 &       &       &       &    &  \cmark  &       &   VGG-16    & \cmark  &    &       &   \cmark     \\
		ATOM~\cite{danelljan2019atom}    & CVPR19 &       &       &       &    &    &       &   Res-18    & \cmark  &    &       &   \cmark     \\
		C-RPN~\cite{fan2019siamese}    & CVPR19 &       &       &       &    &    &       &   AlexNet    &   &    &       &   \cmark     \\
		GFSDCF~\cite{xu2019joint}    & ICCV19 &       &       &       &    &    &       &   Res-50    & \cmark  &    &       &   \cmark     \\
		DiMP~\cite{bhat2019learning}    & ICCV19 &       &       &       &    &    &       &   Res-50    & \cmark  &    &       &   \cmark     \\
		SPLT~\cite{yan2019skimming}    & ICCV19 &       &       &       &    &    &       &   Res-50    & \cmark  &    &       &   \cmark     \\
		GlobalTrack~\cite{huang2019globaltrack}    & AAAI20 &       &       &       &    &    &       &   Res-50    &   &    &       &   \cmark     \\
		LTMU~\cite{dai2020high}    & CVPR20 &       &       &       &    &    &       &   Res-50    & \cmark  &    &       &   \cmark    \\
		\hline
	\end{tabular}%
	\label{tab:summary_tracker}%
\end{table*}%

\noindent {\bf Full Overlap Protocol.} In the full overlap protocol, 1,400 sequences of 70 categories in part-1 are used for training and testing. Specifically, following the 80/20 principle (\ie, the Pareto principle), we select 16 out of 20 sequences in each category for training, and the rest for testing. This way in the full overlap protocol, the training and testing sets consist of 1,120 and 280 videos respectively. Since the number of videos in each category for both training and testing are equal, LaSOT is category-balanced. Table~\ref{tab:full_overlap_training_testing} compares statistics of training/testing sets in full overlap protocol.

\noindent {\bf One-shot Protocol.} In the one-shot protocol, all 1,550 videos from the 85 classes are utilized for training and testing. Because training and testing sets are required to have no overlap in category, we employ 1,400 sequences of 70 categories in part-1 for training, and the other 150 videos of 15 classes in part-2 are used for evaluation. In particular, to increase the source difference, the 15 objects categories are specially chosen outside of the 1,000 classes from ImageNet. It is worth noting that LaSOT is still category-balanced because in both sets, each category contains the same number of videos. Table~\ref{tab:one_shot_training_testing} compares statistics of training/testing sets in one-shot protocol.

\begin{figure*}[!t]
	\centering
	\includegraphics[width=\linewidth]{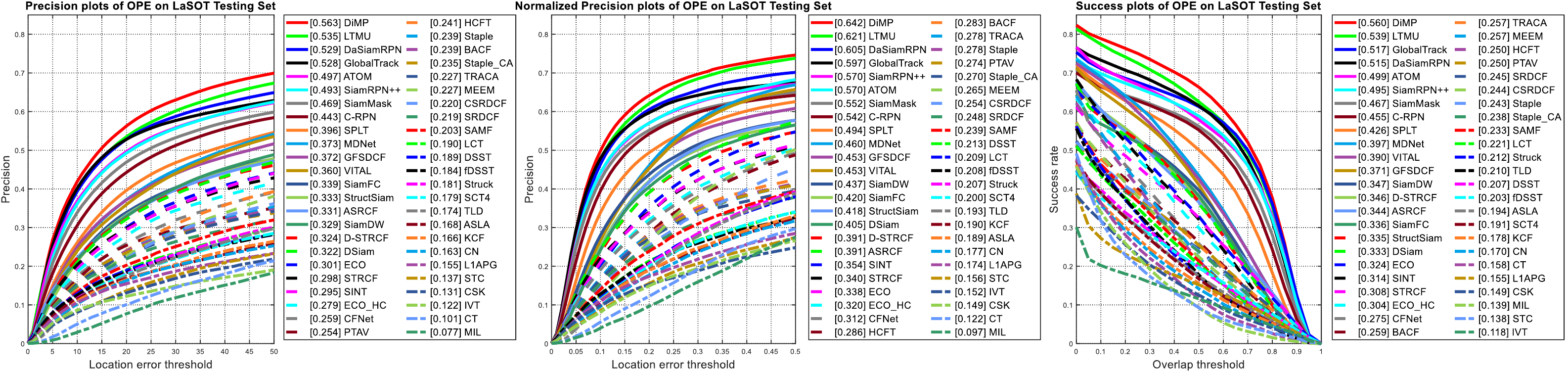}   % full_overlap_overall_new_new
	\caption{Overall evaluation results on LaSOT under the \emph{full overlap} protocol. Best viewed in color.}
	\label{fig:full_overlap_overall_res}
\end{figure*}

\section{Evaluation}

\subsection{Evaluation Metric}

Following~\cite{wu2015object}, we perform One-Pass Evaluation (OPE) and measure the performance of different trackers using three metrics, \ie, {\bf precision}, {\bf normalized precision} and {\bf success}, under two protocols.

The precision (\textbf{PRE}) is calculated by comparing distance between centers of the groundtruth bounding box and the tracking result in pixels. Different algorithms are ranked according to the value of this metric on a certain threshold (\eg, typically 20 pixels). Since PRE does not take object scale into consideration, it is sensitive to target size and image resolution. To avoid this problem, we adopt an additional strategy as in~\cite{muller2018trackingnet} to normalize the PRE with scales. Please refer to~\cite{muller2018trackingnet} for more details. The resulted normalized precision (\textbf{N-PRE}) can ensure the consistency of evaluation across different target scales. The success rate (\textbf{SUC}) is computed as the ratio of the number of successfully tracked frames (\ie, intersection-over-union (IoU) between groundtruth bounding box and tracking result larger than a pre-defined threshold, typically, 0.5) to the number of all frames in a sequence.

\begin{figure*}[!t]
	\centering
	\includegraphics[width=\linewidth]{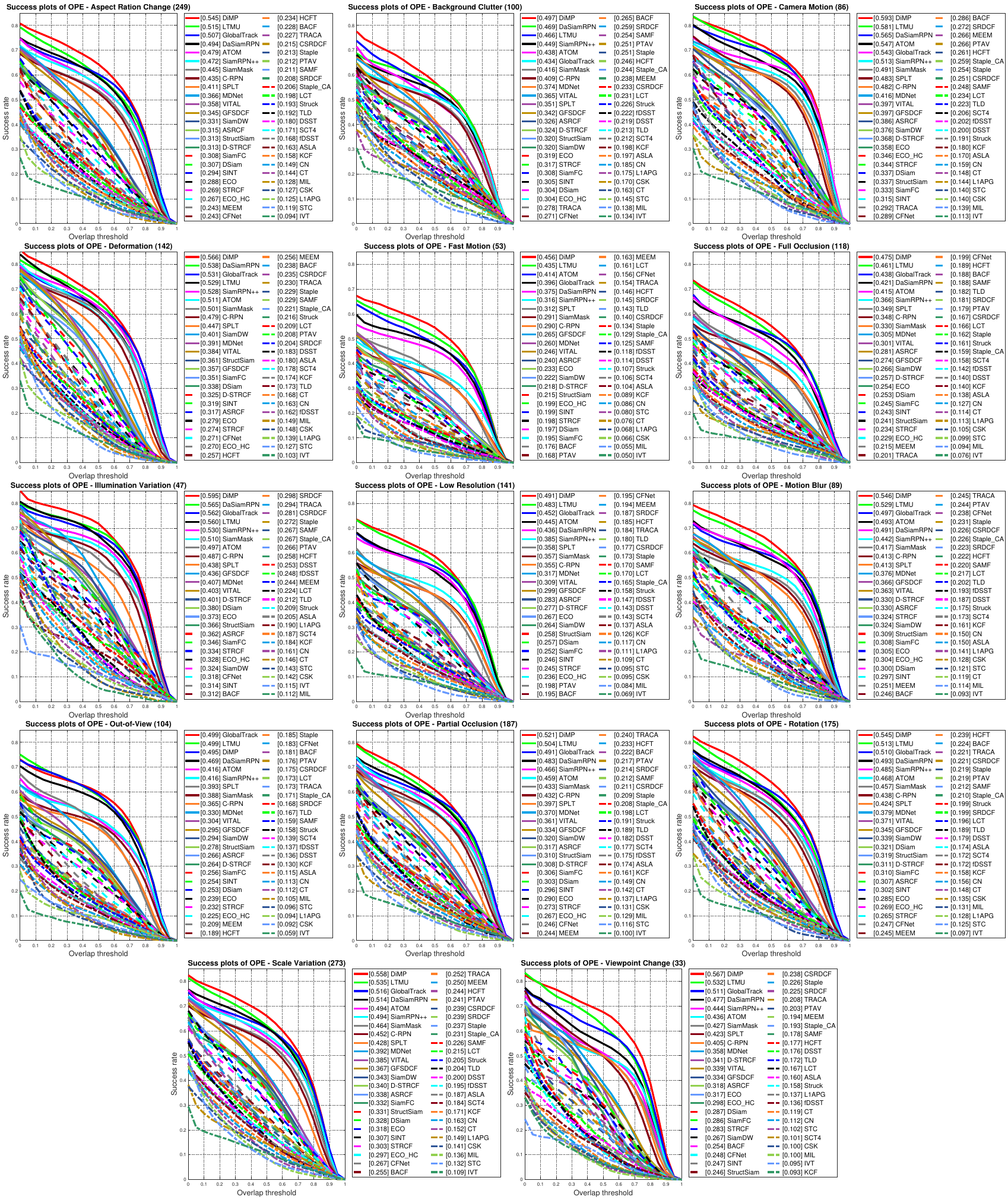} \\
	\caption{Performance of trackers on each attribute using success under \emph{full overlap} protocol. Best viewed in color.}
	\label{fig:full_overlap_all_att_res_success}
\end{figure*}

\begin{figure*}[!t]
	\centering
	\includegraphics[width=\linewidth]{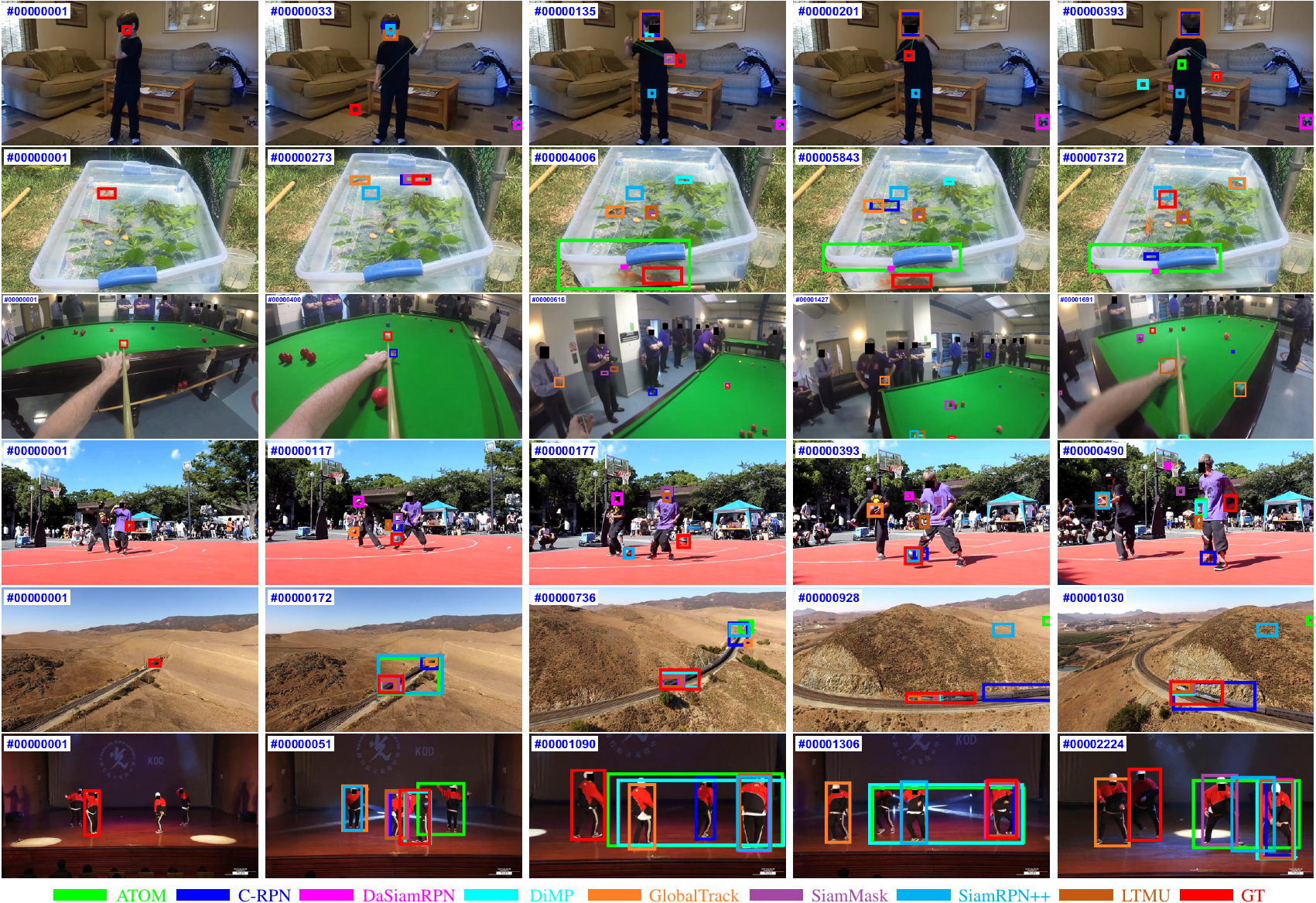}\\
	\caption{Qualitative evaluation in six typically hard challenges in testing sequences of full overlap protocol (from top to bottom): \emph{yoyo-7} with fast motion, \emph{goldfish-7} with full occlusion, \emph{pool-3} with low resolution, \emph{basketball-11} with out-of-view, \emph{train-1} with aspect ration change and \emph{person-5} with background clutter. Best viewed in color.}
	\label{qualitative_sample}
\end{figure*}

\subsection{Evaluated Tracking Algorithms}

In order to provide baselines for future comparison on LaSOT, we extensively evaluate 48 algorithms. In specific, these 48 approaches consist of deep trackers (\eg, MDNet~\cite{nam2016learning}, TRACA~\cite{choi2018context}, CFNet~\cite{valmadre2017end}, SiamFC~\cite{bertinetto2016fully}, StructSiam~\cite{Zhang2018structured}, DSiam~\cite{guo2017learning}, SINT~\cite{tao2016siamese}, ATOM~\cite{danelljan2019atom}, DiMP~\cite{bhat2019learning}, VITAL~\cite{song2018vital}, SiamRPN++~\cite{li2019siamrpn++}, DaSiamRPN~\cite{zhu2018distractor}, \\ SiamDW~\cite{zhang2019deeper}, C-RPN~\cite{fan2019siamese} and SiamMask~\cite{wang2019fast}, GlobalTrack~\cite{huang2019globaltrack}), correlation trackers with hand-crafted features (\eg, ECO\_HC~\cite{danelljan2017eco}, DSST~\cite{danelljan2014accurate}, CN~\cite{danelljan2014adaptive}, CSK~\cite{henriques2012exploiting}, KCF~\cite{henriques2015high}, fDSST~\cite{danelljan2017discriminative}, SAMF~\cite{li2014scale}, SCT4~\cite{choi2016visual}, STC~\cite{zhang2014fast} and Staple~\cite{bertinetto2016staple}) or deep features (\eg, HCFT~\cite{ma2015hierarchical}, D-STRCF~\cite{li2018learning}, ECO~\cite{danelljan2017eco}, GFSDCF~\cite{xu2019joint}, ASRCF~\cite{dai2019visual}) and regularization techniques (\eg, SRDCF~\cite{danelljan2015learning}, STRCF~\cite{li2018learning}, BACF~\cite{galoogahi2017learning}, Staple\_CA~\cite{mueller2017context} and CSRDCF~\cite{lukezic2017discriminative}), ensemble trackers (\eg, SPLT~\cite{yan2019skimming}, LTMU~\cite{dai2020high}, PTAV~\cite{fan2017parallel}, LCT~\cite{ma2015long}, MEEM~\cite{zhang2014meem} and TLD~\cite{kalal2012tracking}), sparse trackers (\eg, L1APG~\cite{bao2012real} and ASLA~\cite{jia2012visual}),  other representatives (\eg, CT~\cite{zhang2012real}, IVT~\cite{ross2008incremental}, MIL~\cite{babenko2009visual} and Struck~\cite{HareST11}). In evaluation, each tracker is used as it is, without any modification. Table~\ref{tab:summary_tracker} summarizes these trackers with their representation schemes and search strategies in a chronological order.

Note that in our evaluation, each tracker is tested as it is in the original paper, for three reasons. First, each tracker may require different training strategy. As a consequence, it is difficult to optimally train all trackers to obtain the best performance. Moreover, inappropriate training settings may result in performance drop for certain trackers. Second, despite using different training data, most deep trackers, especially recently proposed ones, have been fully trained on multiple large scale benchmarks. It is reasonable to assume that each tracker has attained optimal or decent performance in the originally published paper. Third, for trackers that only employ pre-trained classification backbone for feature extraction, it is hard to fine-tune the feature backbone network using existing tracking benchmarks.

\subsection{Evaluation with Full Overlap Protocol}

\subsubsection{Overall performance}

Figure~\ref{fig:full_overlap_overall_res} reports the evaluation results under full overlap protocol in OPE using precision (PRE), normalized precision (N-PRE) and success rate (SUC). DiMP achieves the best performance with PRE score of 0.563, N-PRE score of 0.642 and SUC score of 0.560. DiMP consists of two components including for target localization and scale estimation, both trained on a large set of videos. In addition, the target localization part is online updated during tracking. LTMU shows the second best performance with a 0.535 PRE score, 0.621 N-PRE score and 0.539 SUC score. LTMU focuses on long-term tracking by combining different components such as local tracker and detector. DaSiamRPN obtains the third best results with a 0.529 PRE score, 0.605 N-PRE score and 0.515 SUC score. This method is developed based on SiamRPN++ but utilizes more training data with augmentation techniques. Besides, a re-detection strategy and online model update are adopted for robust long-term tracking. Therefore, DaSiamRPN performs better than its baseline SiamRPN++ with a 0.493 PRE score, 0.57 N-PRE score and 0.495 SUC score. GlobalTrack introduces a two-stage framework for long-term tracking and demonstrates competitive results with a 0.528 PRE score, 0.597 N-PRE score and 0.517 SUC score. ATOM obtains promising results with a 0.497 PRE score, 0.57 N-PRE score and 0.499 SUC score. ATOM introduces an specific network to deal with scale variation. In addition, it employs complex method for optimization and acceleration to achieve real time speed. SiamFC tracker, which learns offline a matching function for tracking, achieves competitive results with a 0.339 PRE score, 0.42 N-PRE score and 0.336 SUC score. It is worth noticing that, unlike the performance on small benchmarks (\eg, OTB-15~\cite{wu2015object}), SiamFC performs better than many more complicated algorithms such as StructSiam, DSiam, PTAV, and HCFT. A possible reason is that these complicated methods are more prone to overfit to small datasets, or they require more hyperparameter tuning to obtain better performance. By contrast, the simple SiamFC has better generalization ability in more challenging and diverse scenarios.

An important observation is that, all top 18 trackers leverage deep features for tracking, which demonstrates the advantages of deep representation in achieving robust tracking performance. Moreover, we observe that model update is beneficial for achieving robust tracking, reflected by superior performance of trackers with online update (\eg, DiMP and LTMU) than those without model update (\eg, GlobalTrack, SiamRPN++ and SiamMask).  

\subsubsection{Attribute-based performance}         

In order to further analyze the performance of different trackers, we conduct attribute-based evaluation.

Figure~\ref{fig:full_overlap_all_att_res_success} shows the attribute-based evaluation results of 48 tracking algorithms with SUC scores under the full overlap protocol. From Figure~\ref{fig:full_overlap_all_att_res_success}, we observe DiMP achieves the best performance under 13 out of 14 attributes. LTMU obtains the second best results under 11 out of 14 attributes. It is worth noting that although the three trackers LTMU, GlobalTrack, and DaSiamRPN utilize additional re-detection strategy for long-term tracking, DiMP still outperforms them under the challenge of occlusion. There are two potential reasons: First, DiMP uses a relatively larger search region for target localization. This way, DiMP can re-locate the target when it reappears. Second, DiMP adopts a more discriminative approach to update the appearance model. Thus, it shows more robust performance when the target re-appears. An interesting observation on out-of-view is that GlobalTrack and LTMU outperform DiMP, which suggests that the full image search strategy is beneficial to handle out-of-view. ATOM obtains promising performance on all attributes owing to the effectiveness of scale estimation networks. In addition, other trackers such as SiamRPN++ and SiamMask achieve competitive results on these 14 attributes. We note that all the top seven trackers, including DiMP, LTMU, GlobalTrack, DaSiamRPN, ATOM, SiamRPN++ and SiamMask, employ deeper feature representation (\eg, ResNet-18 or ResNet-50~\cite{he2016deep}) for appearance modeling, which shows the importance of powerful features for visual tracking.

\subsubsection{Qualitative evaluation}

To qualitatively analyze different trackers and provide guidance for future research, we show sampled tracking results of eight top performers, including DiMP, LTMU, GlobalTrack, DaSiamRPN, ATOM, SiamRPN++, SiamMask and C-RPN, under challenges such as \emph{fast motion}, \emph{full occlusion}, \emph{low resolution}, \emph{out-of-view}, \emph{aspect ratio change} and \emph{background clutter} in Figure~\ref{qualitative_sample}.

From Figure~\ref{qualitative_sample}, we observe that, for sequence \emph{yoyo-7} with \emph{fast motion}, trackers are prone to lose the target because most current algorithms perform target localization from a relatively small region. Although DiMP, LTMU, GlobalTrack, and DaSiamRPN utilize a large search region or adopt re-detection strategies, they still fail as \emph{fast motion} easily causes \emph{motion blur}, which significantly affects re-localization performance of these four trackers. A possible solution to handle this issue is to combine rich temporal and motion cues with appearance information for tracking. In video \emph{goldfish-7} with \emph{full occlusion}, trackers drift to the background region. In order to deal with occlusion, an additional detection component is required to improve performance. All tracking algorithms fail on the video \emph{pool-3} because of the ineffective representation for small target objects. To deal with this, one feasible strategy for deep trackers is to combine multi-scale features from various layers to incorporate details into representation. Video \emph{basketball-11} is difficult due to the \emph{out-of-view} challenge. Similar to the solution for handling \emph{occlusion}, one can leverage an extra instance-level detector to re-locate the target object. \emph{Aspect ratio change} is challenging in \emph{train-1} as most existing trackers often adopt a simple method (\eg, random search or pyramid strategy) to deal with it. A few algorithms such as ATOM and SiamRPN++ borrow techniques from detection for tracking and show promising results. However, since targets may also have \emph{rotation} at the same time, these trackers cannot accurately localize the objects. To effectively estimate target scale, a solution is to take rotation factor into consideration. For video \emph{person-5} with heavy {\em background clutter}, all trackers drift due to less discriminative representation for target and background. A possible solution to alleviate this issue is to utilize the contextual information to enhance the discriminability or fine-grained feature presentation to improve target recognition ability.

\begin{figure*}[!t]
	\centering
	\includegraphics[width=\linewidth]{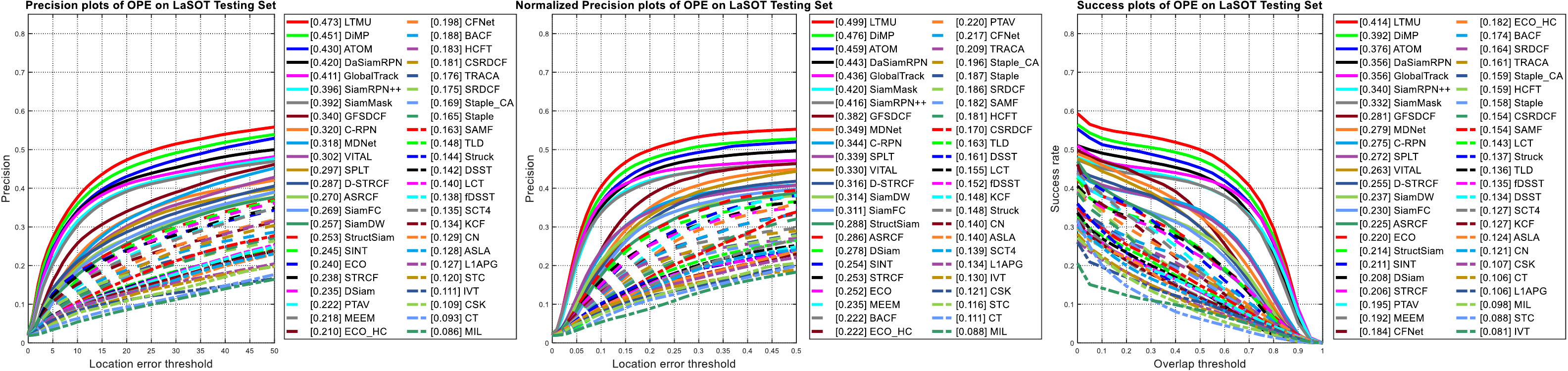}
	\caption{Overall evaluation results on LaSOT under \emph{one-shot} protocol. Best viewed in color.}
	\label{fig:one_shot_overall_res}
\end{figure*}

\subsection{Evaluation with One-Shot Protocol}

\subsubsection{Overall performance}

Different from full overlap protocol, videos for evaluation in the one-shot protocol are from \emph{unseen} categories. In LaSOT, 150 sequences (about 380K frames) from 15 classes are used for performance assessment, and none of the 15 classes is included in the training set or in ImageNet. Figure~\ref{fig:one_shot_overall_res} demonstrates the evaluation results of all algorithms in OPE setting. From Figure~\ref{fig:one_shot_overall_res}, LTMU obtains the best results with a 0.473 PRE score, 0.499 N-PRE score and 0.414 SUC score. DiMP exhibits the second best performance with a 0.451 PRE score, 0.476 N-PRE score and 0.392 SUC score. ATOM achieves the third best results with PRE score of 0.43, N-PRE score of 0.459 and SUC score of 0.376. DiMP performs more robustly than ATOM because it exploits more background information to improve discriminability. We also notice that, in this protocol, ATOM shows better results than DaSiamRPN with a 0.42 PRE score, 0.443 N-PRE score and 0.356 SUC score. This is may be due to their different model update strategies. GlobalTrack obtains promising performance with 0.411 PRE score, 0.436 N-PRE score and 0.356 SUC score. For targets from \emph{unseen} categories, compared to linear template update, supervised online learning can better adapt to appearance changes during tracking, leading to more robust performance. Note that, despite the risk of drifting, online learning trackers usually select reliable tracking results based on their confidences for update. This way, the drift problem can be alleviated to some extent during updating. Similar results have shown that trackers using deeper features such as LTMU, DiMP, ATOM, DaSiamRPN, GlobalTrack, SiamRPN++ and SiamMask achieve better results.

\begin{figure*}[!t]
	\centering
	\includegraphics[width=0.99\linewidth]{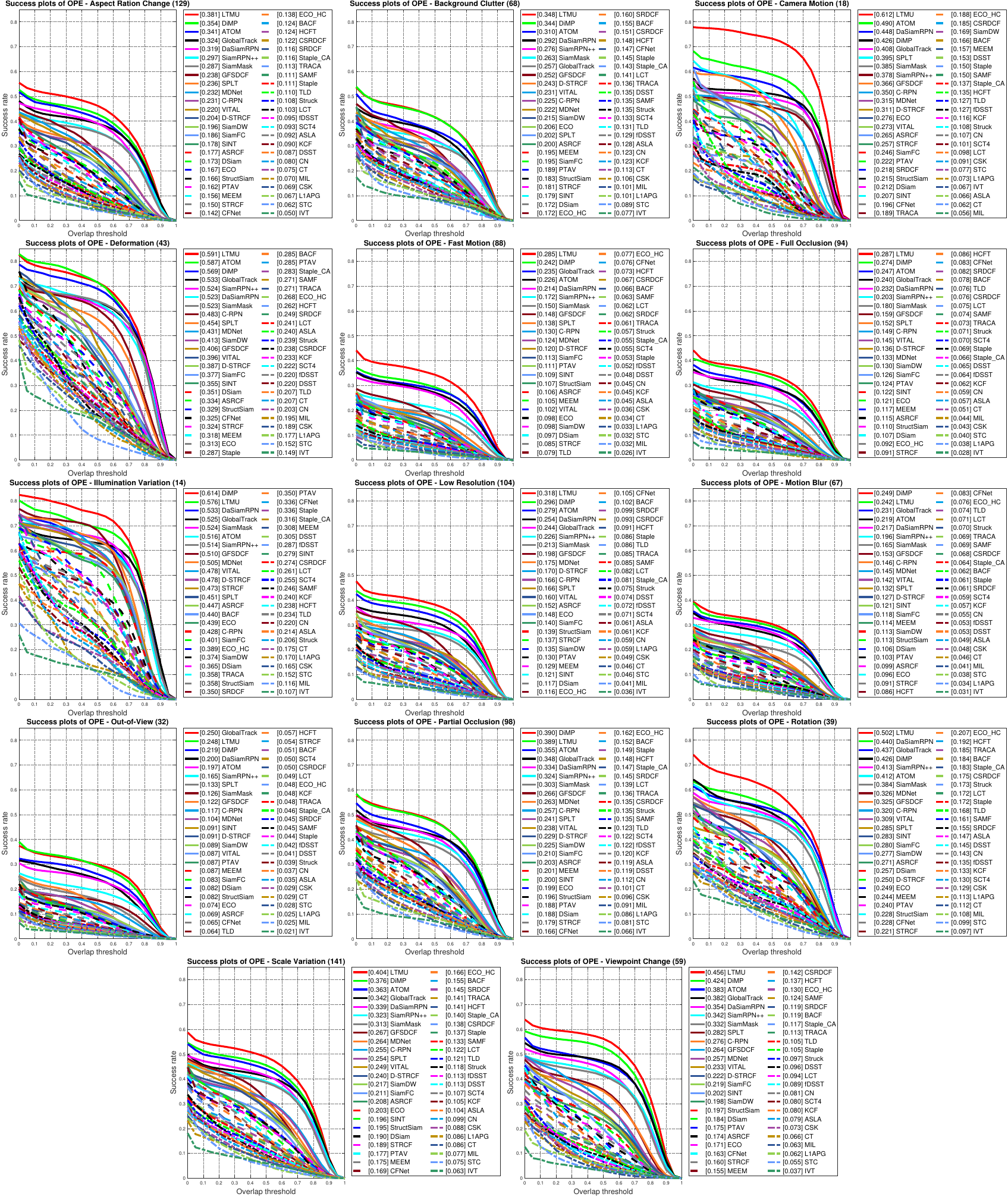}\\
	\caption{Performance of trackers on each attribute using success under \emph{one-shot} protocol. Best viewed in color.}
	\label{fig:one_shot_all_att_res_success}
\end{figure*}

\subsubsection{Attribute-based performance}

Figure~\ref{fig:one_shot_all_att_res_success} demonstrates the attribute-based evaluation results of 45 trackers. We observe that DiMP achieves the best results on 10 out of 14 attributes. DiMP shows the best performance on 3 attributes and the second best results on 8 attributes, demonstrating slightly better performance than GlobalTrack, ATOM and DaSiamRPN. A surprising finding is that despite better overall results of DiMP, GlobalTrack outperforms it on the challenge of out-of-view, thanks to the global search strategy. In addition, an interesting observation is that, SiamMask, which integrates segmentation into tracking for improvement, does not show better performance than DaSiamRPN and SiamRPN++. We conjecture that it is caused by the lack of mask annotation for training SiamMask on our benchmark.

\begin{figure*}[!t]
	\centering
	\includegraphics[width=\linewidth]{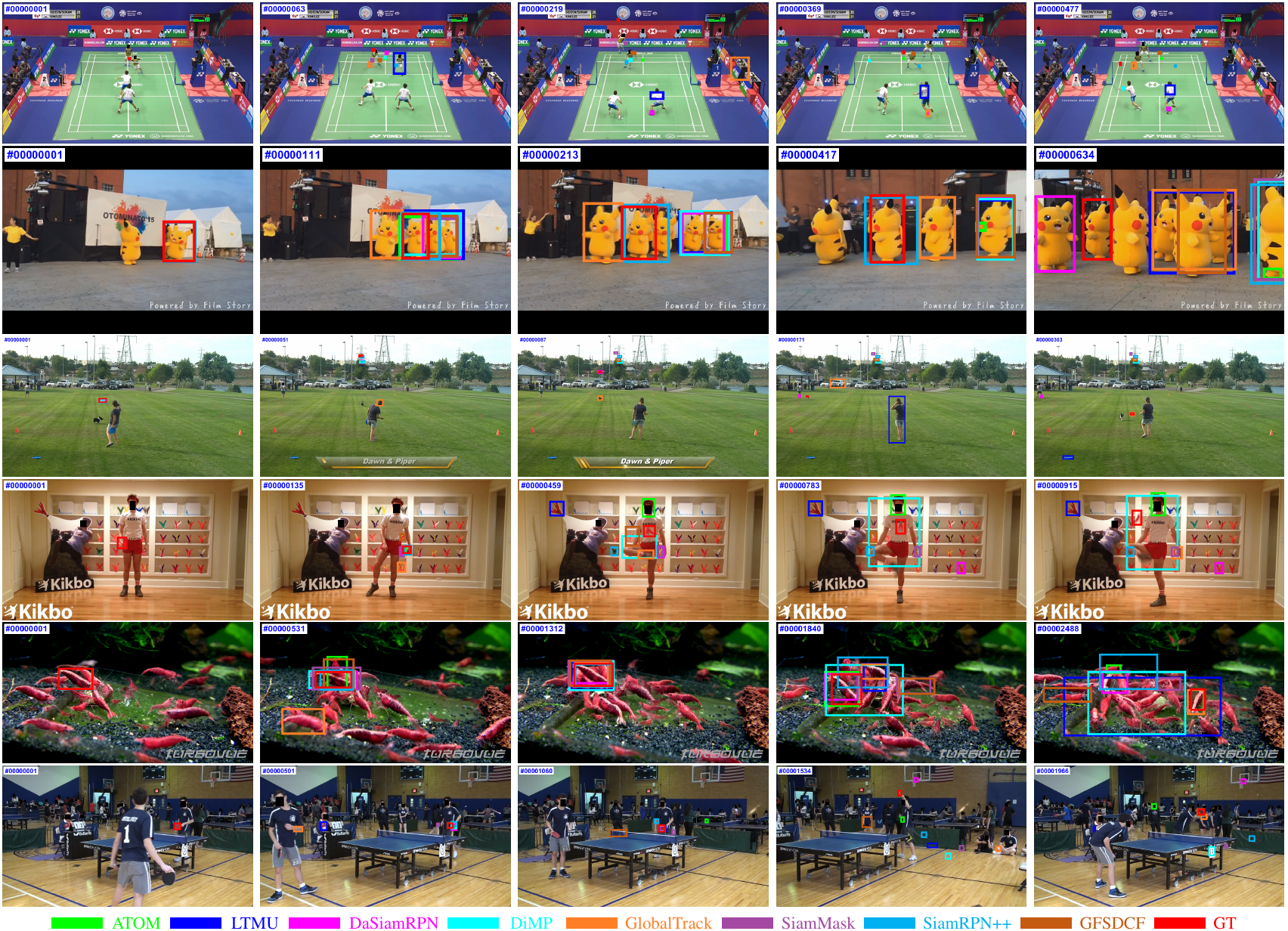}\\
	\caption{Qualitative evaluation in six difficult challenging videos in the one-shot protocol (from top to bottom): \emph{badminton-1} with \textit{fast motion}, \emph{cosplay-8} with \textit{full occlusion}, \emph{frisbee-2} with \textit{low resolution}, \emph{jianzi-4} with \textit{rotation}, \emph{misc-10} with \textit{background clutter} and \emph{paddle-6} with \textit{scale variation}. Best viewed in color.}
	\label{qualitative_sample2}
\end{figure*}

\subsubsection{Qualitative evaluation}

We show qualitative results of eight trackers, including LTMU, DiMP, ATOM, DaSiamRPN, GlobalTrack, SiamRPN++, SiamMask and GFSDCT, in six representative challenges such as \emph{fast motion}, \emph{full occlusion}, \emph{low resolution}, \emph{rotation}, \emph{background} and \emph{scale variation} in Figure~\ref{qualitative_sample2}. For videos with \emph{fast motion} and \emph{full occlusion} (\eg, \emph{badminton-1} and \emph{cosplay-8}), trackers easily drift because they usually utilize a relatively small search for target localization. A solution is to enlarge the search region accordingly or even perform tracking on the full image. For sequences with \emph{low-resolution} and \emph{rotation} (\eg, \emph{frisbee-2} and \emph{jianzi-4}), the tracking algorithms may lose the target because of ineffective feature extraction for target appearance. A feasible method to handle this issue is to mine for motion features in videos. When \emph{background clutter} happens with many distractors (\eg, \emph{misc-10}), it is hard for trackers to locate the target. To solve this issue, one can exploit more spatial details of target to improve discriminative ability of tracking models. In addition, trackers are prone to drift when heavy \emph{scale variation} occurs with other challenges such as \emph{aspect ratio change} (\eg, \emph{paddle-6}). One can leverage techniques such as instance segmentation to improve scale estimation.

\subsection{Retraining on LaSOT}

In order to show the advantages of large-scale training set, we retrain two representative trackers SiamFC~\cite{bertinetto2016fully} and CFNet~\cite{valmadre2017end} using sequences from LaSOT instead of VID for video object detection. Notice that all training settings are kept the same as those for training on VID. After re-training, we compare the performance of these two trackers on different benchmarks including OTB-13, OTB-15, and LaSOT$_{\mathrm{tst}}$ in both protocols. 

Table~\ref{tab:retrain} demonstrates the results of retraining using our dedicated benchmark and comparisons with the performance of the original SiamFC and CFNet trained on ImageNet VID~\cite{russakovsky2015imagenet}. We observe that for both trackers, the performance is improved. Specifically on OTB-13, the SUC score of SiamFC is improved from 0.588 to 0.608 using training split in our full overlap protocol. Furthermore, because of there being more data in the one-shot protocol, the SUC score is increased to 0.614 with significant gains of 2.6\%. On OTB-15, the SUC score of SiamFC is improved from 0.565 to 0.582 and 0.589 with training data from two protocol settings, respectively. Similarly, the SUC score of CFNet obtains obvious improvements on both OTB-13 and OTB-15. More specifically, the SUC score is improved from 0.589 to 0.615 and 0.622 on OTB-13 using training sets in two protocols, and on OTB-15 the score is increased from 0.568 to 0.593 and 0.598. 

\renewcommand\arraystretch{1.25}
\begin{table*}
	\centering
	\caption{Retraining experiments of two trackers SiamFC~\cite{bertinetto2016fully} and CFNet~\cite{valmadre2017end} on LaSOT.}
	\begin{tabular}{@{}r|ccc|ccc@{}}
		\hline
		\multicolumn{1}{c|}{} & \multicolumn{3}{c|}{SiamFC~\cite{bertinetto2016fully}} & \multicolumn{3}{c}{CFNet~\cite{valmadre2017end}} \\
		\cline{2-4} \cline{5-7}
		\multicolumn{1}{c|}{Training set} & \tabincell{c}{ImageNet \\ VID~\cite{russakovsky2015imagenet}}   & \tabincell{c}{LaSOT$_{\mathrm{tra}}$ \\ (full overlap)} & \tabincell{c}{LaSOT$_{\mathrm{tra}}$ \\ (one-shot)} & \tabincell{c}{ImageNet \\ VID~\cite{russakovsky2015imagenet}}   & \tabincell{c}{LaSOT$_{\mathrm{tra}}$ \\ (full overlap)} & \tabincell{c}{LaSOT$_{\mathrm{tra}}$ \\ (one-shot)} \\
		\hline \hline
		OTB-13~\cite{wu2013online} &  0.588 & 0.608 ($\uparrow${\bf 0.020}) & 0.614 ($\uparrow${\bf 0.026}) & 0.589 & 0.615 ($\uparrow${\bf 0.026}) & 0.622 ($\uparrow${\bf 0.033}) \\
		\hline
		OTB-15~\cite{wu2015object} &  0.565 & 0.582 ($\uparrow${\bf 0.017}) & 0.589 ($\uparrow${\bf 0.024}) & 0.568 & 0.593 ($\uparrow${\bf 0.025}) & 0.598 ($\uparrow${\bf 0.030}) \\
		\hline
		LaSOT$_{\mathrm{tst}}$ (full overlap) &  0.336 & 0.342 ($\uparrow${\bf 0.006}) & -     & 0.275 & 0.286 ($\uparrow${\bf 0.011}) & - \\
		\hline
		LaSOT$_{\mathrm{tst}}$ (one-shot) &  0.230  & -     & 0.237 ($\uparrow${\bf 0.007}) & 0.184 & -     & 0.194 ($\uparrow${\bf 0.010}) \\
		\hline
	\end{tabular}%
	\label{tab:retrain}%
\end{table*}%

In addition, we re-evaluate these two trackers on LaSOT under two protocols after retraining, as shown in Table~\ref{tab:retrain}. For SiamFC, the SUC scores under two different protocols are improved from 0.336 to 0.342 and from 0.230 to 0.237, respectively. For CFNet, the SUC scores are improved from 0.275 to 0.286 and from 0.184 to 0.194, respectively. These performance gains show the advantages of large-scale training dataset for improving tracking performance. It is worth noting that, the improvements on the smaller datasets OTB-13 and OTB-15 are higher than those on our testing sets. One possible reason is that our testing sequences are more challenging with focus on long-term tracking, while the original trackers are designed for short-term tracking. For better performance, one may need to adjust more hyperparameters or even design new frameworks.

\section{Discussion}

\subsection{Full Overlap Versus One-Shot}

By definition, the full overlap protocol allows overlap of object classes between training and testing sets, while one-shot protocol does not allow such overlap. Not surprisingly, the one-shot protocol is more challenging because the tracking algorithms need to generalize to objects with unfamiliar appearance and motion pattern.

By comparing the success score of each tracker on the one-shot protocol against the full overlap one, we observe an obvious performance drop (by 0.037-0.18) for all algorithms. Such degradation clearly suggests that existing trackers do not fully address the domain gap between different object categories. To mitigate the performance degradation caused by such domain gap, a potential future direction is to explore domain adaption~\cite{ganin2015unsupervised} for tracking by treating each category or even each target as an individual domain. In addition, by comparing all trackers within full overlap or one-shot protocols, we see that all top five trackers (see Figure 8 and Figure 11) employ deep features for target appearance representation, which shows that designing more effective feature representations should be paid attention to in both scenarios. Considering the dynamic nature of tracking problems, future research can leverage both spatial appearance information and motion features to improve tracking for both seen and unseen object categories. Moreover, we observe that for the top five trackers in each protocol, the best three update the model during tracking, which suggests model updating is critical for both protocols. %We argue that it may play an even more important role in one-shot tracking, because one cannot predict the object's appearance. In such cases, exploiting intermediate status is crucial for learning an effective target model.

\subsection{Short-term and Long-term Tracking Algorithms}

One goal of our benchmark is to advance the development of long-term tracking algorithms. In full overlap evaluation, DiMP achieves the best results and outperforms the long-term tracker LTMU. We argue that the reasons are two-fold. First, DiMP utilizes a relatively large search region for tracking, which effectively handles the problems of full occlusion and out-of-view. Second, the update method in DiMP leverages more historic information than LTMU. In addition, long-term tracker GlobalTrack outperforms ATOM and SiamRPN++ owing to deeper feature representation and a better mechanism to locate target objects using the full image. On the other hand, in one-shot evaluation, LTMU achieves the best performance with SUC score of 0.414. Compared to LTMU, DiMP still achieves competitive results with 0.392 SUC score. The reason that LTMU outperforms DiMP in the one-shot protocol is because there are many small targets. As a result, the tracking model may fail due to ineffective feature extraction and fast target motion. LTMU employs a global search strategy to re-locate the target when drift happens, leading to better results. Moreover, we note that although GlobalTrack adopts full image search methods, its result with 0.356 is inferior in comparison to DiMP, which suggests the importance of effective model updating for robust performance.

Based on the above analysis, we argue that there are several directions that can be taken to improve long-term tracking. First, a deeper feature representation (\eg, ResNet-50) can help to effectively distinguish targets from their backgrounds. Second, a larger search region may be helpful for occluded and out-of-view targets. Third, although matching based trackers (\eg, GlobalTrack and SiamRPN++) achieve promising results in long-term tracking, model updating is still crucial to obtaining more robust performance (\eg, LTMU and DiMP). %We hope that these observations can help facilitate future research on long-term tracking.

\subsection{Analysis on Deeper Feature Representation for Tracking}

Feature representation has been one of the most important components for robust tracking. In this subsection, we conduct experiments by comparing different backbones in both protocols. We choose SiamRPN++ and DiMP for experiments since both approaches provide official implementations with different backbone architectures. Specifically, we study SiamRPN++ with AlexNet, ResNet-18 and ResNet-50 and DiMP with ResNet-18 and ResNet-50. The experimental results are demonstrated in Table~\ref{tab:deep_feat_comparison}.

\renewcommand\arraystretch{1.25}
\begin{table}[!t]\small
	\centering
	\caption{Comparison experiments of different architectures on two protocols using success score.}
	\begin{tabular}{@{}cccc@{}}
		\hline
		& Architectures & Full Overlap & One-shot \\
		\hline
		\hline
		\multirow{3}[0]{*}{SiamRPN++} &AlexNet & 0.433 & 0.245 \\
		& ResNet-18 & 0.472 & 0.316 \\
		& ResNet-50 & 0.495 & 0.340 \\
		\hline
		\multirow{2}[0]{*}{DiMP} & ResNet-18 & 0.534 & 0.381 \\
		& ResNet-50 & 0.560 & 0.392 \\
		\hline
	\end{tabular}%
	\label{tab:deep_feat_comparison}%
\end{table}%

From Table~\ref{tab:deep_feat_comparison}, we can see that on full overlap evaluation, SiamRPN++ with AlexNet achieves a success score of 0.433 and the performance is further improved to 0.472 and 0.495 success scores using deeper architectures ResNet-18 and ResNet-50, respectively. Similarly, DiMP with deeper architecture ResNet-50 shows a better success score of 0.560, outperforming DiMP with ResNet-18 achieving 0.534 success score. Likewise, on one-shot evaluation, SiamRPN++ with deeper ResNet-50 achieves the better performance with a success score of 0.340 compared to the scores of 0.316 and 0.245 achieved with ResNet-18 and AlexNet. DiMP with ResNet-50 obtains a higher score of 0.392 than the score of 0.381 achieved with ResNet-18. The above comparison clearly suggests that feature representation learned by deeper networks demonstrates better robustness for tracking in both full overlap and one-shot protocols. In addition, an interesting observation is that deeper networks are crucial when dealing with unseen targets. When changing backbones from ResNet-18 to AlexNet, the performance drop for SiamRPN++ is 0.039 on the full overlap evaluation. However, on one-shot evaluation, the performance degradation is more obvious with a drop of 0.071 when using AlexNet, which shows that deeper feature representation is more important for tracking performance in locating unseen targets.

\subsection{Analysis on Model Update for Tracking}

Visual tacking is an ill-posed problem in which only information from the first frame is reliable. Due to target appearance variation in video, tracking models usually need an update strategy to handle appearance variation. However, because of occlusion and inaccurate intermediate results, model updating is an extremely complex process. For example, it is difficult to determine when and how to utilize current information for updates. Inappropriate updates may increase the risk of drifting. To avoid this issue, existing trackers such as GlobalTrack, SiamRPN++, SiamMask, and C-RPN formulate tracking as a matching problem without model updates. These approaches show promising performance by achieving success scores of 0.517, 0.495, 0.467 and 0.455 on full overlap evaluation and 0.356, 0.340, 0.332 and 0.275 on one-shot evaluation. In comparison to these trackers without updates, methods with model update including DiMP, LTMU, ATOM, and DaSiamRPN obtain better success scores of 0.560, 0.539, 0.515 and 0.499 on full overlap evaluation and 0.392, 0.414, 0.376 and 0.356 on one-shot evaluation. In addition, we observe that the evaluation of most attributes demonstrates that trackers with model update show better performance. Through the above comparison and analysis, we argue that although online learning for model update is not key to performance improvement, it is essential to perform model updates to achieve robust tracking. We hope that this analysis can inspire future research for better design of tracking algorithms.

\section{Conclusion}

In this paper, we introduced LaSOT, a high-quality large-scale single-object tracking benchmark containing 1,550 videos with more than 3.87 million frames. To our knowledge, LaSOT is by far the \emph{largest} tracking benchmark, in terms of precisely annotated frames. By releasing LaSOT, we expect to offer the community a dedicated platform to develop deep trackers and evaluate long-term tracking performance. In addition, we provided additional lingual specification for each sequence, aiming to encourage the exploration of lingual features to further improve performance. Moreover, for flexible performance evaluation we designed two different experimental settings: the full overlap and one-shot protocols. Extensive experiments on LaSOT by assessing 48 trackers indicate that there is still significant room for future improvement.

\begin{acknowledgements}
We thank the anonymous reviewers for insightful suggestions, and Jeremy Chu for proofreading the final draft. Ling was supported partially by the Amazon AWS Machine Learning Research Award.
\end{acknowledgements}


\bibliographystyle{spmpsci}      % mathematics and physical sciences
\bibliography{reference}
 
%% Non-BibTeX users please use
%\begin{thebibliography}{}
%%
%% and use \bibitem to create references. Consult the Instructions
%% for authors for reference list style.
%\end{thebibliography}

\end{document}